\documentclass[pdflatex,sn-mathphys-num]{sn-jnl}

\usepackage{graphicx}%
\usepackage{multirow}%
\usepackage{amsmath,amssymb,amsfonts}%
\usepackage{amsthm}%
\usepackage{mathrsfs}%
\usepackage[title]{appendix}%
\usepackage{xcolor}%
\usepackage{textcomp}%
\usepackage{manyfoot}%
\usepackage{booktabs}%
\usepackage{algorithm}%
\usepackage{algorithmicx}%
\usepackage{algpseudocode}%
\usepackage{listings}%
\usepackage{booktabs}
\usepackage{tabularx}
\usepackage{pifont}
\usepackage{amssymb}

\usepackage{pifont}
\usepackage{url}

\usepackage{adjustbox}
\usepackage{tabularx}
\usepackage{makecell}
\usepackage{rotating}
\usepackage{nameref}

\theoremstyle{thmstyleone}
\newtheorem{proposition}{Proposition} %
\newtheorem{remark}{Remark}         %

\theoremstyle{thmstyletwo}%

\theoremstyle{thmstylethree}%
\begin{document}

\title[Degeneration of Sliding-Window Factor Graph Optimization into Iterated Extended Kalman Filtering]{Degeneration of Sliding-Window Factor Graph Optimization into Iterated Extended Kalman Filtering}

\author*[1]{\fnm{Baoshan} \sur{Song}}\email{baoshan.song@connect.polyu.hk}

\author[1]{\fnm{Ruijie} \sur{Xu}}\email{ruijie.xu@connect.polyu.hk}

\author[1]{\fnm{Zhi} \sur{Zhan}}\email{25115264r@connect.polyu.hk}

\author*[1]{\fnm{Li-Ta} \sur{Hsu}}\email{lt.hsu@polyu.edu.hk}

\affil*[1]{\orgdiv{Department of Aeronautical and Aviation Engineering}, \orgname{The Hong Kong Polytechnic University}, \orgaddress{\street{Hung Hom}, \city{Hong Kong}, \postcode{999077}, \country{China}}}

\abstract{Sliding window factor graph optimization (SW-FGO) is widely recognized for its robustness, yet its theoretical relationship with the extended Kalman filter (EKF) remains a subject of debate. This paper establishes the sufficient conditions to bridge SW-FGO with the iterated extended Kalman filter (IEKF). We introduce recursive FGO (Re-FGO), a conceptual perspective that employs a two-stage marginalization pipeline to mathematically degenerate the factor graph optimization to the IEKF recursive update. By enforcing the Markov assumption and a single-state window, we prove the theoretical equivalence between the IEKF and Re-FGO. This degeneration is validated through simulations and real-world urban GNSS and INS tightly coupled fusion experiments. The results confirm that Re-FGO exactly reproduces IEKF estimation behavior, demonstrating that the two-stage marginalization pipeline is foundational to enforce structural consistency, thereby successfully uniting graph-based smoothing and filtering paradigms under unified optimization principles.
}


\maketitle

\section{Introduction}\label{sec:introduction}
Wireless technologies, including Wi-Fi \cite{biswas_wifi_2010}, bluetooth \cite{zhuang_bluetooth_2022}, ultra-wideband (UWB) \cite{barbieri_uwb_2021}, 5G \cite{italiano_5g_2023}, and global navigation satellite system (GNSS) \cite{wen_gnss_2021}, are used to provide navigation to various location-based Service (LBS) applications, such as precision agriculture \cite{guo_multi-gnss_2018}, intelligent traffic \cite{tang_deep_learning_2017}, safety, and rescue \cite{atif_uav_2021}. In radio-frequency (RF)-based navigation, the measurements from these wireless technologies are used to estimate states, such as position, velocity and attitude. Often, these measurements are affected by surrounding environments, which might break the commonly used assumption, the non-biased Gaussian distribution, for estimation and filter theories \cite{huber_robust_2011}. A salient example is the bias and noise arisen from multipath effects and non-line-of-sight (NLOS) reception \cite{misra_gps_2002}. These effects challenge the state estimation theories, resulting in unexpected navigation errors. Thus, estimating navigation states from these affected measurements has become a core need, drawing substantial research interest over the past decades.

\begin{figure}
    \centering
    \includegraphics[width=1\linewidth]{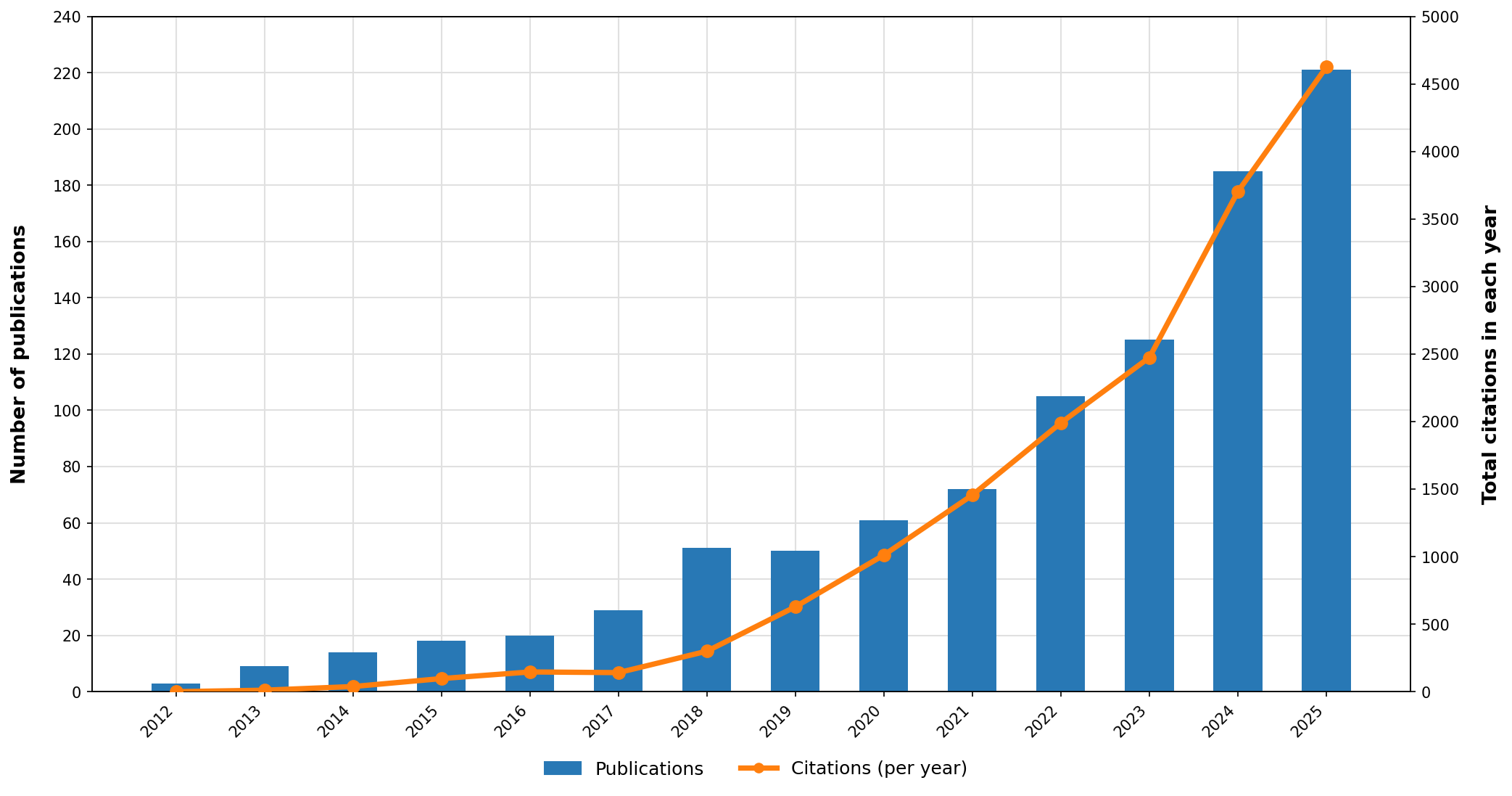}
    \caption{Publications of FGO in navigation fields and their citation from 2012 to 2025. Note that we only collect data from Web of Science with keywords of "graph optimization" "navigation" "localization" or "positioning" in titles. Therefore, many works of simultaneous localization and mapping (SLAM) are not included here.}
    \label{fig:paper_count}
\end{figure}

Traditionally, the state estimation in navigation fields relies on the EKF \cite{kalman_linear_1960,jazwinski_stochastic_2013}. According to  \cite{gelb_applied_1974}, EKF achieves optimal with linear dynamics and measurement models under known white Gaussian noise assumptions. Based on these assumptions, the Markov chain is used to derive the EKF analytical solution (which is rigorously proven to be optimal) \cite{kalman_linear_1960}. The benefit of using the Markov chain is to reduce the computation load, hence enabling EKF to be easily computed in real-time scenarios. With a perfect Markov chain (i.e., no information loss during filtering if everything aligns with the assumptions), EKF can just maintain previous and current state distributions with a two-stage process including prediction and update. Unfortunately, the assumptions on linear and Gaussian models are often challenged when employing sophisticated measurement technologies in complex environments. A prime example is indoor positioning, where multipath effects and NLOS reception frequently violate these idealized conditions \cite{al-jazzar_toa_2007}\cite{marano_nlos_2010}. To relax the assumptions on linear models and Gaussian noises, various Kalman filter variants have been proposed to handle non-Gaussian uncertainties, such as iterated EKF (IEKF) \cite{gelb_applied_1974} and robust EKF (R-EKF) \cite{masreliez_robust_2003}, etc.. Nevertheless, as environmental complexity increases, factor graph optimization (FGO) has emerged as a powerful alternative.

In recent years, FGO, a Bayesian inference framework that represents states and measurements as nodes in a bipartite graph, has gained popularity in navigation community due to its effectiveness in solving nonlinear problems with non-Gaussian noises. Before its moment in navigation, the notion of factor graph was popular in the computer science field \cite{kschischang_factor_2002}\cite{loeliger_introduction_2004}. Fig. \ref{fig:paper_count} shows the rising interest in FGO-based methods for navigation. 
Among these works, most of them focus on differences between FGO and EKF at an application level. A particular example is the Google Smartphone Decimeter Challenge (GSDC) \cite{fu_android_2020}. In the first and second challenges (2021-2022), FGO-based methods won the first place award by introducing more constraints and sliding window based estimation to facilitate the effect of GNSS measurement outliers \cite{suzuki_smartphone_2021}\cite{suzuki_2022_1st}. In the 2023 challenge, a refined EKF smoothing framework outperformed other complex models, proving that classic filters remain highly competitive when properly configured \cite{motooka_2024_first}. 
These results from GSDC challenges have exposed the enthusiasm of the community to compare FGO and EKF based methods. However, to dive into the analysis of the results, a fundamental ambiguity remains unresolved. Specifically, the theoretical boundary between EKF and FGO is often blurred in practice. 
Under this ambiguity, it remains impossible to discern whether performance gains stem from the optimization strategy itself or from specific parameter configurations like window size and iteration count. Thus, to construct a state estimator effectively, it is crucial to understand the theoretical relationship between FGO and EKF, and clarify the assumptions under which they become equivalent.

To clarify this ambiguity, several theoretical analyses have explored the relationship. Some typical works are listed as below.
In the navigation community, EKF is first compared theoretically with batch-FGO, which shows the ability of FGO in dealing with non-Gaussian noise effect \cite{li-ta_hsu_navigation_2022}. After that, a tutorial for FGO in navigation derives the formulation of FGO including real-time sliding window based FGO and batch one to solve EKF-based problems \cite{taylor_factor_2024}. Technically, the EKF is often viewed as a limiting case of FGO with a sliding window of size one and a single iteration. Although theoretical equivalence is often assumed, the concrete realization of a window-size-one FGO that exactly matches the EKF is not a straightforward derivation. Mostly the existing SW-FGO framework relies on Schur complement-based marginalization, which focuses on preserving the sparsity of the graph and its Hessian matrix. However, it remains unclear how the SW-FGO framework can be mathematically reconciled with the recursive, two-stage (predict-update) pipeline of EKF. 
Recently, Weng et al. \cite{weng_receding_2025} provided a theoretical proof of equivalence between EKF and moving horizon estimation (MHE), highlighting the critical role of the linearization point. 
While MHE emphasizes time-domain linear approximations, FGO leverages the topological structure of factor graphs to encode probabilistic constraints. How to explicitly represent the Markovian dependency and the sequential relinearization of EKF within the node-and-factor language of FGO is still an open question. Moreover, existing analysis primarily addresses the non-iterative case, leaving the connection between IEKF and FGO underexplored. As both IEKF and FGO feature iterative structures for real-time estimation, unifying their theoretical foundations is highly beneficial.

\begin{figure}
    \centering
    \includegraphics[width=0.7\linewidth]{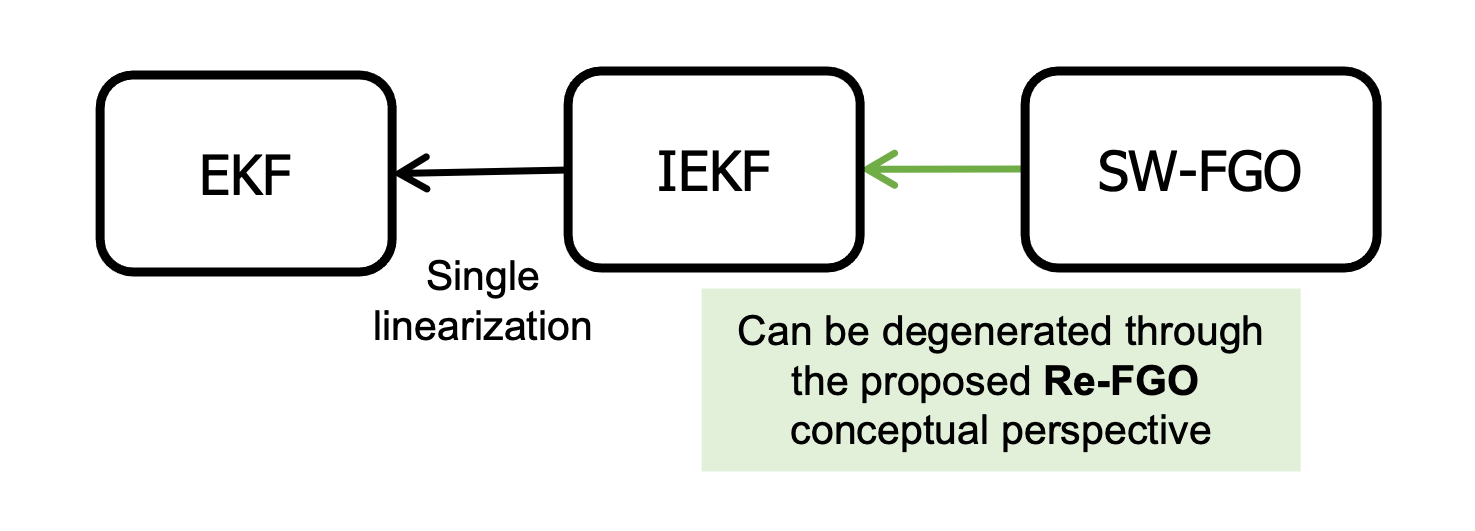}
    \caption{The transformation pipeline between EKF, IEKF and SW-FGO. The flowchart illustrates the transition including under certain conditions. Note that SW-FGO could be degenerated to IEKF based on the proposed Re-FGO conceptual perspective.
} \label{fig:theoretical}
\end{figure}

As FGO has been more and more popular, we believe that it is meaningful to analyze the theoretical relationship between EKF and FGO to explain the results in various applications. 
To make the comparison fair, both the Markov assumption and iteration need to be considered. Hereby, the objective of this paper is to answer the question on how IEKF and FGO could be bridged equivalently. 
Since there are many references for FGO and EKF theory, this paper is not a review or a tutorial but a supplement for those literature.
The objective of this paper is not to propose a new estimator, but to establish a possible perspective to connect IEKF and SW-FGO. 
To achieve this objective, our work focuses on clarifying the mathematical evolution from recursive filtering to graph based optimization. As illustrated in Fig. \ref{fig:theoretical}, we introduce Re-FGO as a sufficient bridge serving as a conceptual perspective that allows the IEKF to be interpreted through the lens of factor graph optimization, thereby revealing how IEKF emerges as degenerated cases of SW-FGO under specific constraints. 

Contributions of this work are four-fold:
\begin{itemize}
    \item First, we establish a topological mapping between directed probabilistic graphical models and undirected factor graphs, providing a unified framework to represent traditionally uni-directed Kalman filtering processes within a graph optimization infrastructure.
    
    \item Second, we formalize a two-stage marginalization pipeline comprising sequential state elimination via Schur complement and factor-only anchoring via QR decomposition, which successfully preserves Markovian consistency and prevents future relinearization drift.
    
    \item Third, we develop a conceptual perspective Re-FGO to implement a sufficient way by which SW-FGO could degenerate to IEKF. This allows for a granular analysis of the EKF-FGO transition chain.
    
    \item Fourth, we conduct simulated and real-world experiments to prove the connection between IEKF and SW-FGO using Re-FGO. At the same time, sensitive tests are also conducted within two-stage marginalization to analyze how iteration and marginalization affect the linearization point in both nonlinear and non-Gaussian cases.
\end{itemize}To the best of our knowledge, this paper explicitly formulates the degeneration process from SW-FGO to IEKF and demonstrates its practical implementation in the context of navigation.

The remainder of this paper is organized as follows. Section \nameref{sec:method} introduces the corresponding methodology of the comparison, including discussion of current research gap, introduction of two-stage marginalization, and the proposed Re-FGO perspective. Section \nameref{sec:evaluation} evaluates the comparison between EKF, IEKF and SW-FGO by simulation and real-world GNSS/INS integration.

\section{Methods} \label{sec:method}

\begin{table}[htbp]
\centering
\caption{Comparison with related works to highlight our contribution.}
\footnotesize 
\begin{tabularx}{\columnwidth}{@{} l *{3}{>{\centering\arraybackslash}X} >{\raggedright\arraybackslash}p{3.2cm} @{}}
\toprule
Reference & Filters & Methods & Evaluation & Contribution \\ \midrule
Barfoot \cite{barfoot_state_2024} & EKF/IEKF & MAP & Simulation & MAP-based derivation \\ \addlinespace[2pt]
Wen \cite{wen_factor_2021} & EKF & SW/batch-FGO & GNSS/INS & Integration comparison \\ \addlinespace[2pt]
Taylor \cite{taylor_factor_2024} & EKF & SW/batch-FGO & Simulation & EKF vs. FGO analysis \\ \addlinespace[2pt]
Weng \cite{weng_receding_2025} & EKF & MHE & Simulation/GNSS & Bridge EKF and MHE \\ \addlinespace[2pt]
\textbf{Ours} & \textbf{EKF/IEKF} & \textbf{SW-FGO} & \textbf{Simulation and GNSS/INS} & \textbf{Degenerate SW-FGO to IEKF} \\ \bottomrule
\end{tabularx}
\label{tab:review}
\end{table}



In this section, as shown in Table \ref{tab:review} and Table \ref{tab:compare_ekf_swfgo}, we analyze the theoretical relationship between EKF and SW-FGO. 
The relationship between recursive filtering and FGO has been profoundly established from a Maximum A Posteriori (MAP) estimation. As elucidated in pioneering works \cite{barfoot_state_2024, taylor_factor_2024}, an EKF can be viewed as a specialized instance of a SW-FGO estimator. Specifically, under the assumptions of a window size $w=1$, a single Gauss-Newton iteration, and first-order Taylor expansion, the two paradigms could yield a similar derivation from the MAP origin. The MAP estimate $\mathcal{X}^*$ is obtained by minimizing the sum of squared mahalanobis distances of the residuals:
\begin{equation} \label{equ:map_appendix}
\mathcal{X}^* = \arg\min_{\mathcal{X}} \sum_{k} \left( \|\mathbf{r}_{f,k}\|^2_{\mathbf{Q}} + \|\mathbf{r}_{h,k}\|^2_{\mathbf{R}} \right)
\end{equation}
where $\mathcal{X}$ denotes the state sequence to be estimated, $\mathbf{r}_{f,k}$ and $\mathbf{r}_{h,k}$ represent the residuals of the motion and measurement models at epoch $k$, while $\mathbf{Q}$ and $\mathbf{R}$ are the covariance matrices representing process and measurement noise levels, respectively. For readers' convenience, a detailed revisiting of the EKF and FGO derivations from a MAP estimation is provided in Supplementary Note 1. 

\subsection{Why SW-FGO with $w=1$ is non-trivial }

\begin{figure}
    \centering
    \includegraphics[width=1\linewidth]{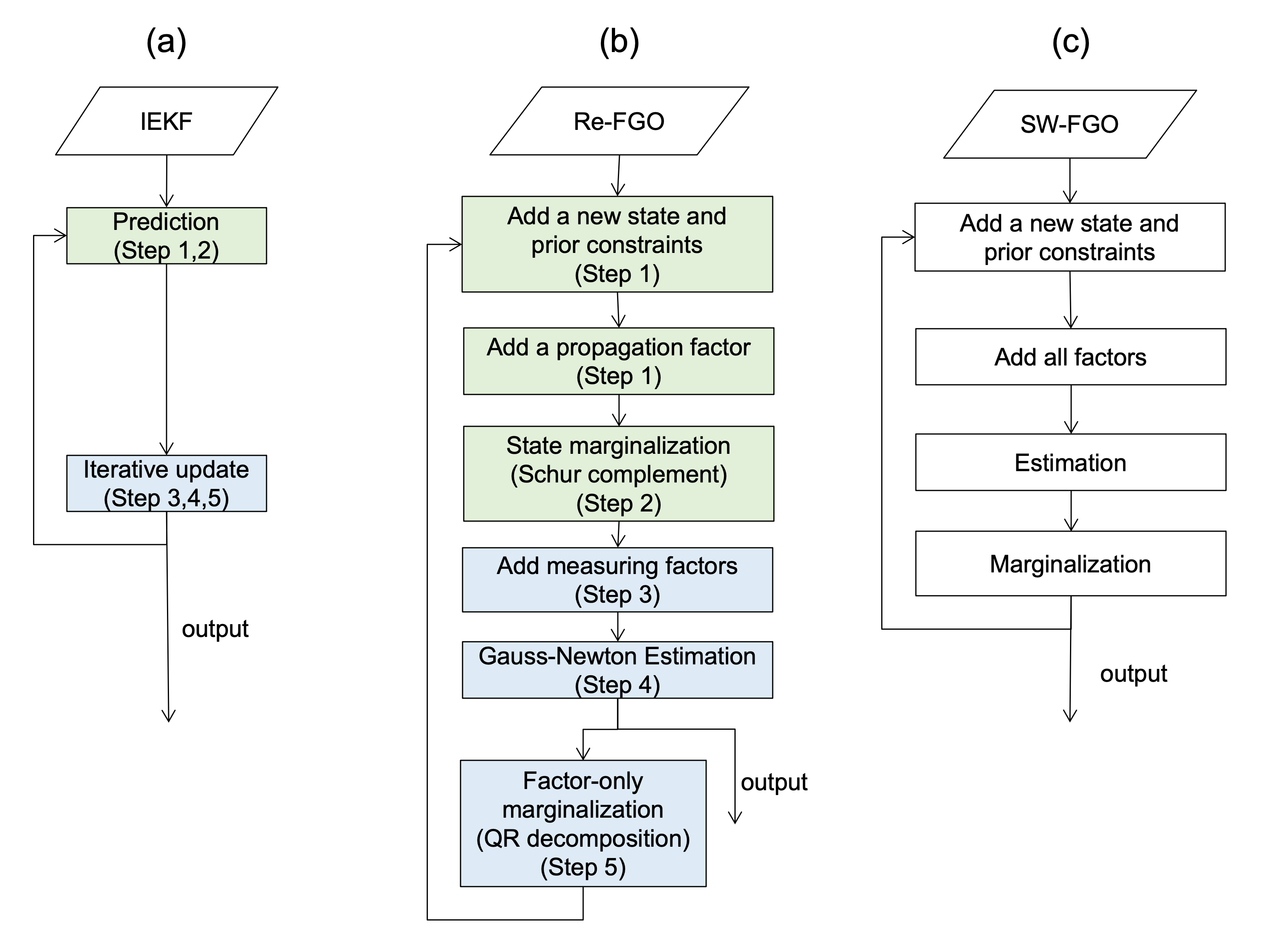}
    \caption{Pipeline comparison of IEKF, Re-FGO, and SW-FGO. (a) IEKF, (b) Re-FGO, and (c) SW-FGO. Green and blue boxes represent the equivalent prediction and update stages across the IEKF and Re-FGO pipelines.}
    \label{fig:pipeline}
\end{figure}

While the macro equivalence between the EKF and SW-FGO is well recognized, a significant divergence persists in their practical realization. Despite their shared Bayesian origin in MAP estimation, EKF and SW-FGO employ fundamentally different operational strategies to resolve the underlying optimization problem.
To the best of our knowledge, a strict derivation and implementation that bridges EKF and a $w=1$ SW-FGO remains elusive, primarily due to two structural barriers.
First, a disparity exists in the implementation of SW-FGO with window of size 1.
The general version of SW-FGO only perform estimation and marginalization once, while EKF adopts Markov assumption and solve the MAP problem with two stages, including prediction and update. Thus, the initialization estimation of each methods could be different. For example, the initial guess including both state value and its uncertainty of the update stage in EKF is given by the prediction stage. However, the initialization of SW-FGO comes from the user input at the first epoch or marginalization at the last epoch. Second, it is difficult to implement the SW-FGO with window of 1. Traditionally, SW-FGO operates as an incremental process where state estimation and marginalization are performed in a single-stage cycle. As shown in Fig. \ref{fig:pipeline}(c), in common sliding window implementations (e.g., $w \ge 2$), a new state is added, the graph is optimized, and the oldest state is marginalized to form a prior. However, this single-stage logic does not naturally degenerate into the $w=1$ structure of the EKF. The most popular marginalization using Schur complement requires at least two variables in the graph:

\begin{equation}
\label{equ:schur_appendix}
\boldsymbol{\Lambda}_{k}^- = \boldsymbol{\Lambda}_{22} - \boldsymbol{\Lambda}_{21} \boldsymbol{\Lambda}_{11}^{-1} \boldsymbol{\Lambda}_{12} = (\mathbf{F}_{k-1} \mathbf{P}_{k-1}^+ \mathbf{F}_{k-1}^T + \mathbf{Q})^{-1}
\end{equation}
In Eq. \ref{equ:schur_appendix}, $\boldsymbol{\Lambda}_{k}^-$ is the marginalized predictive information matrix at epoch $k$, $\boldsymbol{\Lambda}_{ij}$ are the block components of the joint Hessian matrix, $\mathbf{F}_{k-1}$ is the Jacobian of the motion model at $k-1$, and $\mathbf{P}_{k-1}^+$ is the posterior covariance from the previous epoch.

If we need to keep the sliding window size as one, it is unsuitable to perform the Schur complement after estimation. If one attempts a naive $w=1$ SW-FGO by performing a single marginalization at the end of each epoch, the estimator suffers from a structural gap. Therefore, it is non-trivial to find the conditions under which the EKF will be equivalent to SW-FGO. 

Based on the prior discussion, the core distinction between the EKF and SW-FGO lies in the topological handling of information flow. From the perspective of probabilistic graphical models \cite{barfoot_state_2024}, the EKF is inherently a directed graph where information flows with strict temporal causality, propagating unidirectionally from the posterior state at $k-1$ to the state at $k$ via the motion model. Conversely, SW-FGO utilizes an undirected graph language where all states and observations within a window are co-optimized simultaneously. This functional misalignment underpins why simply setting the window size to one fails to reproduce standard filtering behavior. To formally bridge these two disparate formalisms, a new operational paradigm that reshapes the undirected topology dynamically is strictly required.

\begin{table}[]
\caption{Comparison between EKF, IEKF and SW-FGO.}
\begin{tabular}{@{}cccc@{}}
\toprule
 &
  EKF &
  IEKF &
  SW-FGO \\ \midrule
Markov   chain &
  \ding{51} &
  \ding{51} &
  \begin{tabular}[c]{@{}c@{}}Can be employed based on \\the proposed Re-FGO conceptual perspective \\(the main contribution of this paper)\end{tabular} \\
Multiple   linearization &
  Once &
  \ding{51} &
  \ding{51} \\ \bottomrule
\end{tabular} \label{tab:compare_ekf_swfgo}
\end{table}

\begin{figure}
    \centering
    \includegraphics[width=1\linewidth]{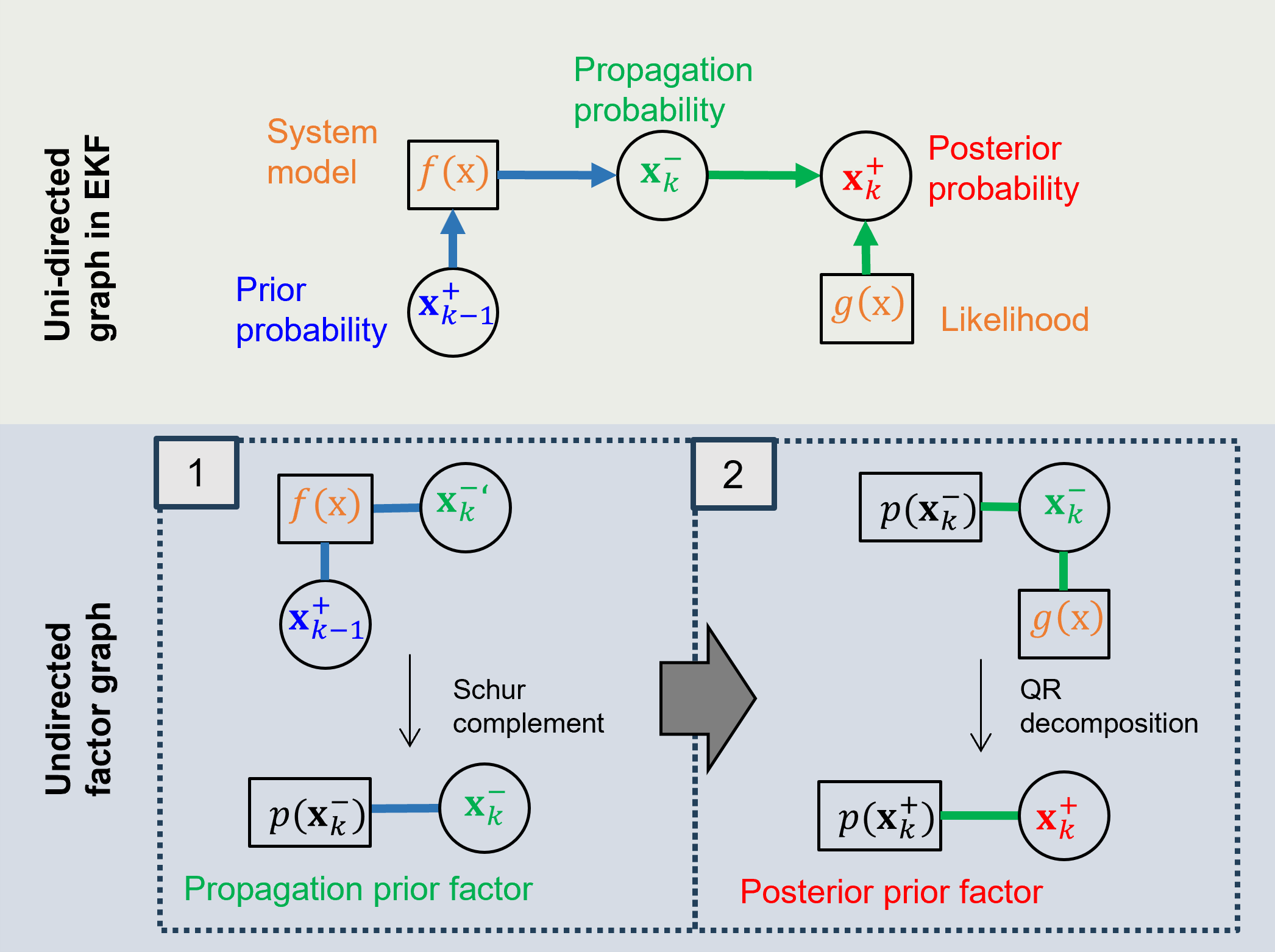}
    \caption{Undirected factor graph representation of uni-directed EKF via two-stage marginalization. Left-down stage 1 (state marginalization): State elimination for prediction. Right-down stage 2 (factor-only marginalization): Factor anchoring for posterior update.}
    \label{fig:two_stage_marg}
\end{figure}

\begin{table}[htbp]
\centering
\small
\caption{Five-Step Correspondence between IEKF and Re-FGO. In the Re-FGO framework, $\mathbf{A}$ and $\mathbf{b}$ represent the joint Jacobian and residual vector of the prior-measurement system; $\mathbf{Q}$ and $\mathbf{R}_{qr}$ denote the orthogonal matrix and the upper-triangular information factor from the Thin QR factorization; $\mathbf{d}$ and $\mathbf{e}$ are the transformed residuals associated with the state update and the orthogonal error respectively. For the increments, $\Delta\mathbf{x} = \mathbf{x}_k^- - \mathbf{x}_k^*$ and $\Delta\mathbf{z} = \mathbf{z}_k - h(\mathbf{x}_k^*)$.}
\label{tab:algorithm_comparison}
\renewcommand{\arraystretch}{2.2} 
\setlength{\tabcolsep}{4pt}      
\begin{tabularx}{\textwidth}{l|X|X}
\toprule
\textbf{Step} & \textbf{IEKF} & \textbf{Re-FGO} \\ \midrule
\textbf{1. State Pred.} & 
$\mathbf{x}_k^- = \mathbf{f}_k(\mathbf{x}_{k-1}^+)$ & 
$\mathbf{x}_k^- = \mathbf{f}_k(\mathbf{x}_{k-1}^+)$ \\ \midrule
\textbf{2. Cov. Pred.} & 
$\boldsymbol{\Lambda}_{k}^{-} = (\mathbf{F}_{k-1} \mathbf{P}_{k-1}^+ \mathbf{F}_{k-1}^T + \mathbf{Q})^{-1} = (\mathbf{P}_k^-)^{-1}$ \\ \midrule
\textbf{3. Iteration} & 
$\mathbf{H}_k^{(i)}, \mathbf{K}_k^{(i)}, \mathbf{x}_k^{(i)}$ & 
$\mathbf{H}_k^{(i)}, \mathbf{A}^{(i)}\delta\mathbf{x} \approx \mathbf{b}^{(i)},$ \newline $\mathbf{x}_k^{(i)} \leftarrow \mathbf{x}_k^{(i-1)}+\delta\mathbf{x}$ \\ \midrule
\textbf{4. Final Upd.} & 
$\delta\mathbf{x}^* = \mathbf{K}_k\Delta\mathbf{z} + (\mathbf{I}-\mathbf{K}_k\mathbf{H}_k)\Delta\mathbf{x}$ \newline
$\mathbf{x}_k^+ = \mathbf{x}_k^* + \delta\mathbf{x}^*$ & 
$\mathbf{A} = \begin{bmatrix} (\mathbf{P}_k^-)^{-1/2} \\ \mathbf{R}_k^{-1/2}\mathbf{H}_k \end{bmatrix}, $\newline
$\mathbf{b} = \begin{bmatrix} (\mathbf{P}_k^-)^{-1/2}\Delta\mathbf{x} \\ \mathbf{R}_k^{-1/2}\Delta\mathbf{z} \end{bmatrix}$ \newline
$\mathbf{Q}^\top\mathbf{A} = \begin{bmatrix}\mathbf{R}_{qr} \\ \mathbf{0}\end{bmatrix}, 
\mathbf{Q}^\top\mathbf{b} = \begin{bmatrix}\mathbf{d} \\ \mathbf{e}\end{bmatrix}$ \newline
$\mathbf{R}_{qr}\delta\mathbf{x}^* = \mathbf{d}, \mathbf{x}_k^+ = \mathbf{x}_k^* + \delta\mathbf{x}^*$ \\ \midrule
\textbf{5. Cov. Upd.} & 
$(\mathbf{P}_k^+)^{-1} = (\mathbf{P}_k^-)^{-1} + \mathbf{H}_k^\top\mathbf{R}_k^{-1}\mathbf{H}_k$ & 
$\mathbf{R}_{qr}^\top\mathbf{R}_{qr} = (\mathbf{P}_k^+)^{-1}$ \\ \bottomrule
\end{tabularx}
\end{table}

\subsection{The Proposed Two-Stage Marginalization Pipeline} \label{sec:two_stage_marginzalization}
To resolve the structural and topological gaps identified above, this section formalizes the exact mathematical and topological operations of the proposed two-stage marginalization pipeline, which is illustrated in Fig. \ref{fig:two_stage_marg}.

\noindent\textbf{Stage 1: State Marginalization (Prediction Equivalent).} The first stage structurally replicates the EKF prediction step by executing variable elimination immediately after a new state $x_k$ and its associated propagation factor are introduced, but {before} the latest measurement is processed. In this step, the historical variable node $x_{k-1}$ is eliminated to establish a predictive prior for $x_k$, compressing the historical information into a predictive prior factor. This operation simulates the directional initial-guess transfer of a directed model and ensures the sliding window is maintained at a minimal size of exactly one.

To trace this topological compression, we focus on the joint Hessian matrix $\boldsymbol{\Lambda}$ of the active factor graph. At this timestamp, the graph consists strictly of the prior information from the previous epoch and the newly introduced propagation factor. By isolating these components, the joint system can be explicitly partitioned with respect to the state dimensions of the historical node $x_{k-1}$ and the incoming node $x_k$:
\begin{equation} \label{equ:partition_stage1}
\boldsymbol{\Lambda} = 
\begin{bmatrix} 
\boldsymbol{\Lambda}_{x_{k-1} x_{k-1}} & \boldsymbol{\Lambda}_{x_{k-1} x_k} \\ 
\boldsymbol{\Lambda}_{x_k x_{k-1}} & \boldsymbol{\Lambda}_{x_k x_k} 
\end{bmatrix} 
\end{equation}
By executing variable elimination on the block components of Eq. \eqref{equ:partition_stage1} via the Schur complement operation derived in Eq. \eqref{equ:schur_appendix}, the historical information from $x_{k-1}$ is collapsed into the dense predictive prior factor $\boldsymbol{\Lambda}_{k}^{-}$ on the surviving node $x_k$:
\begin{equation} \label{equ:schur_pure}
\boldsymbol{\Lambda}_{k}^{-} = \boldsymbol{\Lambda}_{x_k x_k} - \boldsymbol{\Lambda}_{x_k x_{k-1}} \boldsymbol{\Lambda}_{x_{k-1} x_{k-1}}^{-1} \boldsymbol{\Lambda}_{x_{k-1} x_k}
\end{equation}
This operational sequence guarantees that the directional prior transfer of a directed model is strictly simulated, keeping the active sliding window at a minimal size of exactly one state before entering the measurement update.

\noindent\textbf{Stage 2: Factor-Only Marginalization (Update Equivalent).} The second stage functions as the equivalent of the filter's update stage. Once iterative optimization converges to the posterior mode $x_k^*$, a factor-only marginalization is performed to anchor the linear approximation of the active measurement models. Concretely, let $\mathbf{A}$ be the joint system Jacobian stacked from the predictive prior $\boldsymbol{\Lambda}_{k}^{-}$ and the active measurement residuals evaluated at the finalized mode $x_k^*$, and $\mathbf{b}$ the corresponding residual vector. The measurement factor is condensed via QR decomposition \textit{without} removing the state node $x_k$:
\begin{equation}\label{equ:qr}
  \mathbf{Q}^{\top}\mathbf{A}
  = \begin{bmatrix} \mathbf{R}_{qr} \\ \mathbf{0} \end{bmatrix},
  \qquad
  \mathbf{Q}^{\top}\mathbf{b}
  = \begin{bmatrix} \mathbf{d} \\ \mathbf{e} \end{bmatrix},
\end{equation}
where $\mathbf{Q}$ is an orthogonal matrix and $\mathbf{R}_{qr}$ is the upper-triangular information factor. The upper block $(\mathbf{R}_{qr},\,\mathbf{d})$ encodes an {anchored} linearization at $x_k^*$, preventing future re-linearization of historic measurement models. Note that while QR decomposition is standard in the FGO solver GTSAM to manage the Bayes tree \cite{kaess_isam2_2011}, here it serves the structurally distinct purpose of locking the linearization point rather than performing variable elimination. 

Together, these two stages constitute the complete two-stage marginalization pipeline. Only by executing both distinct steps can an undirected factor graph sufficiently replicate the EKF's two-step handling of state estimates and uncertainty without violating the underlying Markov chain topology.
\begin{remark}[State vs. Factor-Only Elimination]\label{remark:elimination}
The two-stage marginalization pipeline relies on two distinct algebraic mechanisms. State elimination, executed via the Schur complement in Stage 1, is a topological operation that marginalizes out a variable node (e.g., $x_{k-1}$) and collapses its incident edges into a dense prior on the surviving node, reducing the state dimensionality. Conversely, factor-only elimination, executed via QR decomposition in Stage 2, targets the measurement model rather than the variable itself. Instead of removing the state node $x_k$, it anchors its linearization point to the current posterior mode. This dual design ensures that temporal priors are propagated while strictly satisfying Markovian consistency and preventing future re-linearization drift.
\end{remark}

\subsection{Degenerate SW-FGO to IEKF via Re-FGO} \label{sec:bridging}
To bridge the structural and topological gaps identified in the preceding sections, we introduce Re-FGO, a conceptual perspective that explicitly enforces the Markov assumption within a factor graph architecture.
As shown in Fig. \ref{fig:pipeline}(b), Re-FGO is a conceptual framework designed to enforce the Markov assumption within an undirected factor graph structure. By setting the window size to $w=1$, the lifecycle of a single epoch $k$ is divided into two sequential optimization and marginalization stages.

\noindent\textbf{Algebraic Formulation of the Two Stages.} In Stage 1, after the propagation factor from $x_{k-1}$ to $x_k$ is introduced but before the current measurement $z_k$ is integrated, Re-FGO eliminates the previous state $x_{k-1}$. This step is governed by the following localized optimization problem:
\begin{equation} \label{equ:re_fgo_stage1}
\arg\min_{x_{k-1}, x_k} \left( \left\| x_{k}-f(x_{k-1})\right\|^2_{\mathbf{Q}^{}} + \left\| x_{k-1}-{x}^+_{k-1}\right\|^2_{\mathbf{P}_{k-1}^+} \right)
\end{equation}
Linearizing the motion model $f(x_{k-1})$ around the prior mode $x_{k-1}^+$ yields the joint system Hessian. By applying the Schur complement to eliminate $\delta x_{k-1}$ as shown in Eq. \eqref{equ:schur_appendix}, we obtain the marginalized predictive information matrix $\boldsymbol{\Lambda}_{k}^{-}$:
\begin{equation} \label{equ:schur_pure}
\boldsymbol{\Lambda}_{k}^{-} = (\mathbf{F}_{k-1} \mathbf{P}_{k-1}^+ \mathbf{F}_{k-1}^T + \mathbf{Q})^{-1} = (\mathbf{P}_k^-)^{-1}
\end{equation}
Equation \eqref{equ:schur_pure} shows that Stage 1 is algebraically identical to the IEKF prediction step, outputting the predictive mean $x_k^- = f(x_{k-1}^+)$ and its information weight $\boldsymbol{\Lambda}_{k}^{-}$.

In Stage 2, the raw measurement $z_k$ is added to the graph. Re-FGO resolves the measurement non-linearities by executing iterative Gauss-Newton updates. At epoch $k$ and iteration $j$, the state update increment $\delta x_{k,j}^*$ is resolved via the following cost function:
\begin{equation} \label{equ:re_fgo_stage2}
\delta x_{k,j}^* = \arg\min_{\delta x_{k,j}} \left( \left\| \mathbf{H}_{k,j} \delta x_{k,j} - \mathbf{r}_{k,j} \right\|^2_{\mathbf{R}^{}} + \left\| \delta x_{k,j} + (x_{k,j} - x_k^-) \right\|^2_{\mathbf{P}_{k}^{-}} \right)
\end{equation}
where $x_{k,j}$ is the current linearization point, $\mathbf{r}_{k,j} = z_k - h(x_{k,j})$ is the measurement residual, and $\mathbf{H}_{k,j}$ is the measurement Jacobian. 
Crucially, the iterative optimization in Eq. \eqref{equ:re_fgo_stage2} establishes the exact numeric bridge to the filter's update phase. Upon convergence to the posterior mode $x_k^*$, the state increment vanishes ($\delta x_{k,j}^* \rightarrow 0$), and the joint system Hessian $\mathbf{H}_{fgo}$ expands to include the finalized measurement Jacobian $\mathbf{H}_k$:
\begin{equation} \label{equ:hessian_expansion}
\mathbf{H}_{fgo} = \boldsymbol{\Lambda}_{k}^{-} + \mathbf{H}_k^T \mathbf{R}^{-1} \mathbf{H}_k
\end{equation}
By substituting the Stage 1 state marginalization outcome (Eq. \eqref{equ:schur_pure}) into Eq. \eqref{equ:hessian_expansion}, the graph's total information metrics can be directly rewritten as:
\begin{equation} \label{equ:equivalence_bridge}
\mathbf{H}_{fgo} = (\mathbf{P}_k^-)^{-1} + \mathbf{H}_k^T \mathbf{R}^{-1} \mathbf{H}_k = (\mathbf{P}_k^+)^{-1}
\end{equation}
Equation \eqref{equ:equivalence_bridge} confirms that the total information matrix of Re-FGO at the converged MAP mode is algebraically identical to the IEKF posterior information form. 

To complete the recursive lifecycle without expanding the graph topology, the finalized joint Jacobian $\mathbf{A}$ is processed via the thin QR factorization formalized in Eq. \eqref{equ:qr}. This step serves as the irreversible condensation of information. By storing exclusively the upper-triangular factor $\mathbf{R}_{qr}$ and the conditioned residual $\mathbf{d}$ as the incoming prior for the subsequent epoch, Re-FGO permanently anchors the linearization point at $x_k^*$. This discrete operational flow effectively blocks the retrospective propagation of future optimization updates, ensuring the estimation sequence adheres strictly to the directional causal structure of a Markov chain.

\noindent\textbf{The Degeneration Proposition.} The sequential operations detailed above allow us to state the explicit conditions under which an undirected graph optimization matches a recursive filter:

\begin{proposition}[Degeneration of SW-FGO to IEKF] \label{prop:relationship_kfv}
Suppose the following four structural conditions hold:
\begin{enumerate}
  \item \textbf{Markov Chain Topology:} The state sequence satisfies the first-order Markov property, meaning the joint probability distribution factorizes sequentially over time.
  \item \textbf{Single-State Window:} The active sliding window is strictly restricted to a size of $w=1$ by executing temporal state marginalization (Stage 1, Eq. \eqref{equ:schur_pure}) at the start of each epoch.
  \item \textbf{Iterative Numerical Equivalence:} Non-linear measurement models are minimized using iterative Gauss-Newton refinement (Eq. \eqref{equ:re_fgo_stage2}) until numerical convergence is achieved within the epoch.
  \item \textbf{Fixed Linearization Anchor:} Upon convergence to the posterior mode $x_k^*$, the measurement Jacobian $\mathbf{H}_{k}$ is permanently locked via factor-only marginalization (Stage 2, Eq. \eqref{equ:qr}).
\end{enumerate}
Under these four conditions, the undirected factor graph optimization exactly replicates the state estimates and covariance updates of the closed-form IEKF.
\end{proposition}

We map the step-by-step mathematical correspondences between the IEKF and the Re-FGO pipeline in Table \ref{tab:algorithm_comparison}. We compare our framework specifically to the IEKF rather than the standard EKF. This choice is mandatory because both SW-FGO and the IEKF use iterative numerical steps to resolve measurement non-linearities, whereas the standard EKF uses a single-step Taylor expansion. By aligning the iterative mechanics, we isolate the structural effects of the marginalization protocols from linearization truncation errors. Finally, a detailed Re-FGO derivation is provided in Supplementary Note 2. 

\begin{remark}[Comparison with MSCKF as a Bridge]\label{remark:msckf}
The Multi-State Constraint Kalman Filter (MSCKF) \cite{clement_msckf_2015} also combines filtering and optimization concepts, but it does not serve as a direct theoretical bridge between standard SW-FGO and the EKF. While the standard EKF condenses all historical data into a single-state prior, the MSCKF maintains a short historical buffer of poses to form multi-state feature constraints. Because it delays marginalization across a sliding window of past states, the MSCKF does not enforce the strict single-state Markov property within each individual epoch. In contrast, Re-FGO maintains a strict window size of $w=1$ via its two-stage marginalization pipeline, isolating how the local Markov assumption alters the consistency of standard factor graphs.
\end{remark}

\section{Results and Discussion}\label{sec:evaluation}
To empirically validate the relationship between SW-FGO and KFV, we conduct both simulation and real GNSS/INS experimental analysis. The core of this evaluation is to verify the Re-FGO perspective through which SW-FGO could degenerate into IEKF, and to analyze the role of the Markov assumption in this process. All experiments are constructed in MATLAB 2024b on an Intel NUC desktop (i7-1260P, 16 GB RAM).

\begin{table}[ht]
\small
\centering
\caption{Four data schemes, where ``L'' and ``NL'' denote linear and nonlinear; ``G'' and ``NG'' denote Gaussian and non-Gaussian distributions; ``GMM'' denotes Gaussian mixture model; ``N(a,b)'' denotes a Gaussian distribution with mean $a$ and standard deviation (std) $b$; anchor radius denotes the radius of the circle where emitters are uniformly placed.}
\label{tab:data_scheme}
\begin{tabularx}{\textwidth}{@{} X l l l l @{}}
\toprule
\textbf{Data Schemes} & \textbf{L+G} & \textbf{NL+G} & \textbf{L+NG} & \textbf{NL+NG} \\ 
\midrule
Noise distribution & G & G & GMM & GMM \\
\addlinespace
Parameters & $N(0, 0.1)$ m & $N(0, 0.1)$ m & \makecell[c]{$0.8 \cdot N(0, 0.1)$ m \\ $+ 0.2 \cdot N(0, 10)$ m}  & \makecell[c]{$0.8 \cdot N(0, 0.1)$ m \\ $+ 0.2 \cdot N(0, 10)$ m}  \\
\addlinespace
Anchor radius & 1000 m & 105 m & 1000 m & 105 m \\ 
\bottomrule
\end{tabularx}
\end{table}

\subsection{Simulation}
The evaluation begins by 
a series of synthetic experiments featuring a moving receiver following a uniform circular motion model, supplemented by time of arrival (TOA) measurements. By modulating the nonlinearity of the TOA model and the measurement noise profile, we generated four distinct datasets to address diverse estimation challenges: L+G (low nonlinearity with Gaussian noise), NL+G (high nonlinearity with Gaussian noise), L+NG (low nonlinearity with non-Gaussian noise), and NL+NG (high nonlinearity with non-Gaussian noise).
 
Based on the Re-FGO perspective, we implement its variant employing only a single linearization per epoch, using the subscript ``1'', denoted as Re-FGO$_{1}$.
Key simulation parameters are detailed in Table \ref{tab:data_scheme}. This setup provides a sufficient basis to prove the numerical equivalence between Re-FGO and IEKF, while highlighting the critical role of the two-stage marginalization in anchoring linearization points.

\begin{figure}
    \centering
    \includegraphics[width=1\linewidth]{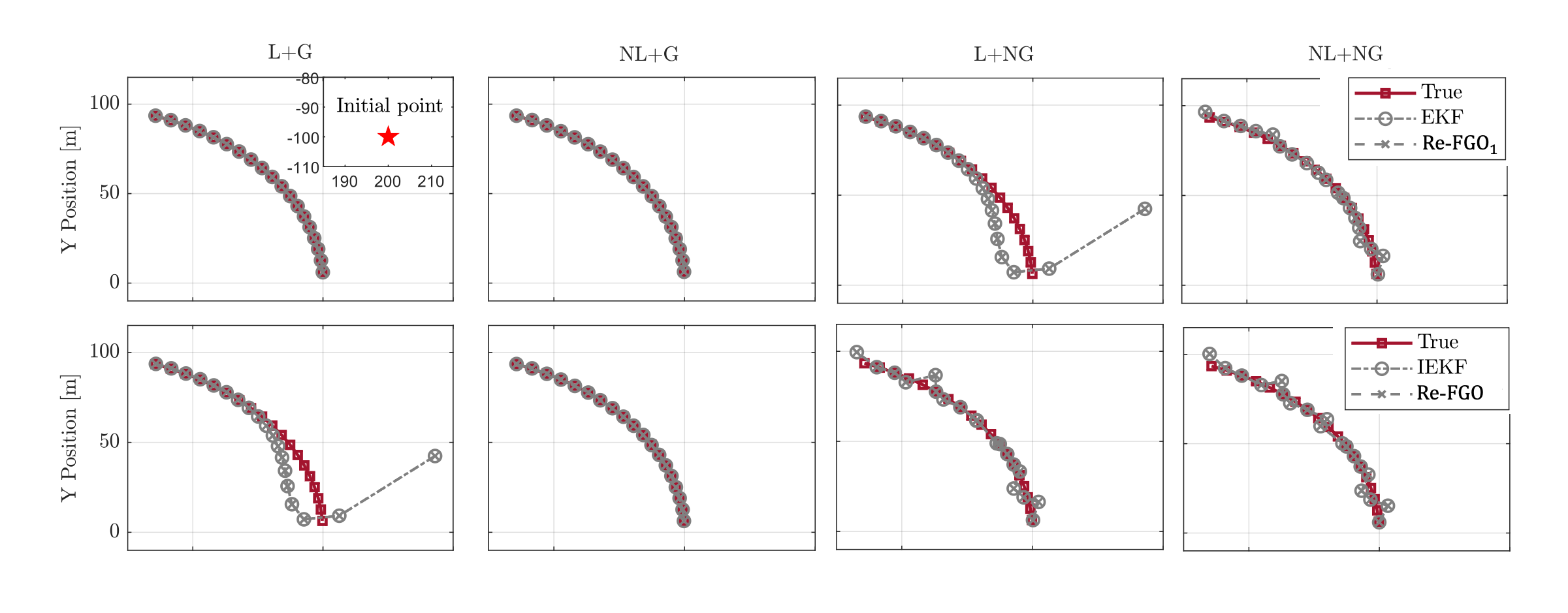}
    \caption{Trajectories comparison between EKF/Re-FGO$_1$ and IEKF/Re-FGO. Results are shown for the first row (EKF/Re-FGO$_1$) and second row (IEKF/Re-FGO) in four groups of TOA/UCM tightly coupled navigation evaluation. Note that there are 100 epochs in total and only 15 epochs are shown for demonstration.}
    \label{fig:final}
\end{figure}

\begin{figure}
    \centering
    \includegraphics[width=0.5\linewidth]{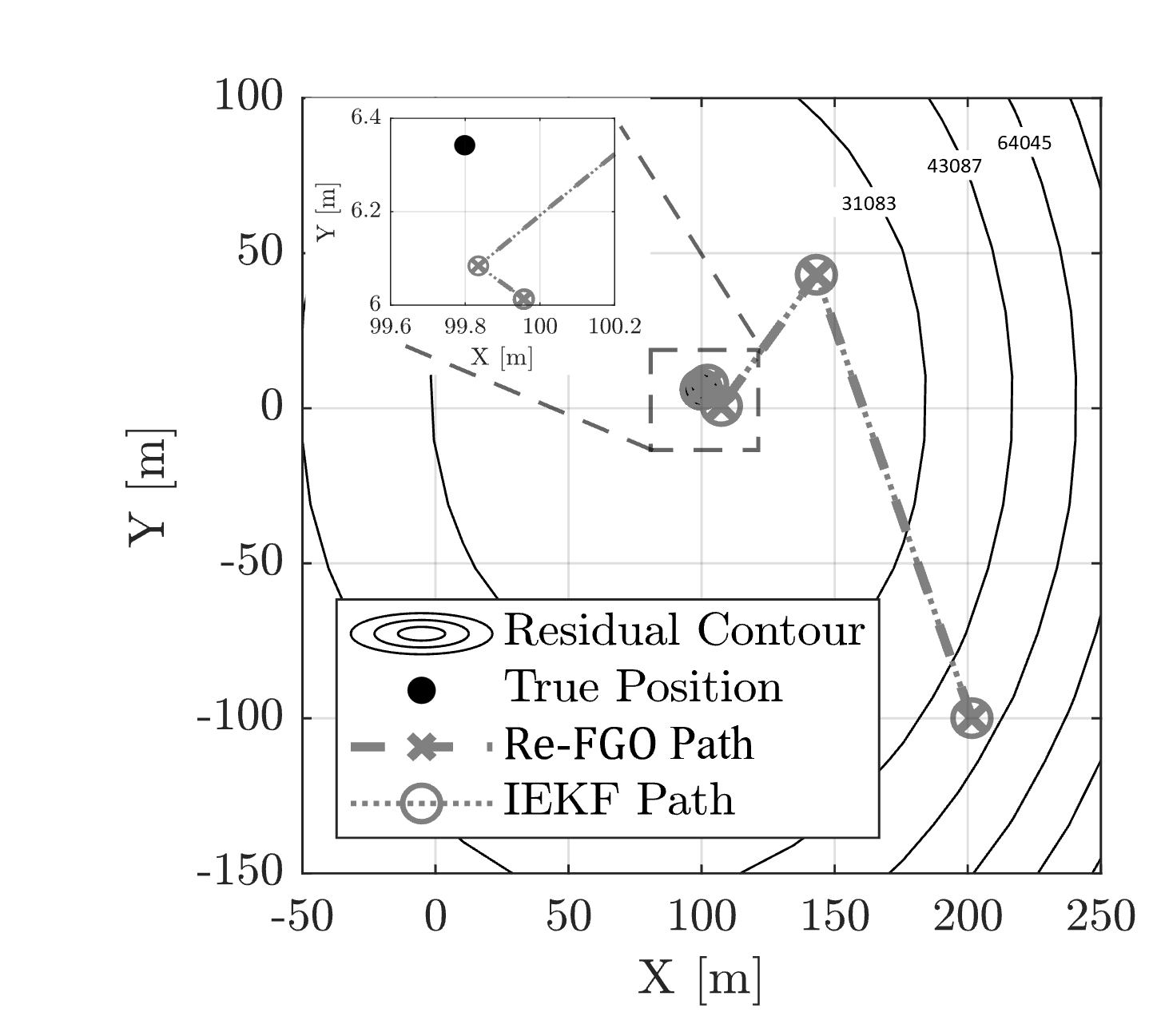}
    \caption{Residual norm descending paths at the first epoch of the NL+NG test. Evaluation with large initial positioning error, comparing IEKF vs. Re-FGO, where the residual contour represents the level of the residual 2-norm.}
    \label{fig:residual_norm_path}
\end{figure}

\begin{table}[]
\caption{Mean trajectory differences between the comparison groups.}
\begin{tabular}{@{}lllll@{}}
\toprule
Mean trajectory difference [m] & L+G      & NL+G     & L+NG     & NL+NG    \\ \midrule
EKF   and Re-FGO$_1$                          & 1.36e-11 & 2.24e-11 & 1.36e-12 & 2.21e-11 \\
IEKF   and Re-FGO                   & 8.17e-14 & 8.88e-15 & 1.19e-13 & 1.02e-14 \\ \bottomrule
\end{tabular}
\label{tab:kfv_fgkfv_auto_err_diff}
\end{table}

\begin{table}[]
\centering
\caption{Comparison of CP95 (95\% confidence probability) positioning errors and absolute percentage differences between Re-FGO and the variant without second marginalization.}
\begin{tabular}{lcccc}
\toprule
CP95 [m] & L+G & NL+G & L+NG & NL+NG \\ 
\midrule
Re-FGO & 0.1717 & 0.1689 & 9.8168 & 9.0481 \\
Re-FGO without second marginalization & 0.1725 & 0.1679 & 9.9116 & 8.9000 \\ 
Absolute difference percentage & 0.47\% & 0.59\% & 0.97\% & 1.64\% \\ 
\bottomrule
\end{tabular}
\label{tab:sw-fgo}
\end{table}

\noindent\textbf{Equivalent results between IEKF and SW-FGO.} Fig. \ref{fig:final} presents a comprehensive 2×4 grid comparing trajectories across various configurations. Each column corresponds to a distinct simulation scenario, while the rows are organized into comparative pairs, specifically EKF vs. Re-FGO$_1$ and IEKF vs. Re-FGO, facilitating a direct validation of the recursive optimization equivalence. 
The most striking observation from Fig. \ref{fig:final} is the exact alignment between each group of comparison. Across all schemes, from the baseline L+G to the extreme NL+NG case, the Re-FGO consistently replicate the positioning performance of IEKF, while Re-FGO$_1$ aligns perfectly with the EKF. 
This identity is quantitatively proved by the error analysis in Table \ref{tab:kfv_fgkfv_auto_err_diff}, where the discrepancies between the two result sets remain at the level of machine precision (approximately $10^{-15}$ to $10^{-11}$ m). Such numerical isomorphism confirms that Re-FGO is not merely a functional approximation but a mathematically identical representation of IEKF within the factor graph domain, provided the Markov assumption and two-stage marginalization are strictly maintained. This equivalence is further elucidated by the residual norm paths in Fig. \ref{fig:residual_norm_path}. The identical residual decay patterns and Jacobian directions expose a common optimization landscape shared by all IEKF and Re-FGO.

Beyond validating equivalence, Fig. \ref{fig:final} highlights the critical role of the linearization point in estimation accuracy. In the heavily nonlinear NL+G and NL+NG scenarios, IEKF significantly outperforms EKF by iteratively re-anchoring the Jacobian at the posterior mode, effectively minimizing geometric bias during the update stage. It is crucial to note that the precision of the linearization point is not only important for internal update iterations but is also fundamental to maintaining the consistency of the Markov chain during the two-stage marginalization process. As will be demonstrated in the following setion, omitting the second marginalization, which essentially create a ``floating'' linearization point, could lead to a misalignment of the recursive information flow. We empirically prove this loss of stability using the Re-FGO variant without the second marginalization.

\noindent\textbf{The Two-stage marginalization for initialization.} While iterative refinement effectively enhances estimation within a single epoch, it only addresses the utilization of instantaneous information. For recursive filters, long-term stability depends on maintaining the consistency of the Markov chain through two-stage marginalization in Section \nameref{sec:two_stage_marginzalization} and Fig. \ref{fig:two_stage_marg}. The first stage mirrors the Schur complement in SW-FGO, where the old state and its associated factor is marginalized to form a temporal prior. In contrast, the second stage is specifically responsible for anchoring the linearization point of the current measurement factors. Given that the most significant nonlinear and non-Gaussian effects in a system typically originate from the measurement models, the accuracy and stability of this linearization point are paramount. Consequently, this experiment employs a sensitive test to compare performance with and without the second stage marginalization, effectively evaluating whether a ``floating'' Jacobian undermines the recursive chain's integrity. Here the ``floating'' means that the Jacobian could be changed by iteration or other estimator behaviors.

As shown in Table \ref{tab:sw-fgo}, our comparative experiments employing the Re-FGO and its variant without second marginalization across four simulation scenarios reveal the effect caused by the failure of anchoring the Jacobian. In the nearly linear L+G scenario, the impact is negligible as the Jacobian remains relatively constant. However, in the NL+G scenario, omitting the second marginalization allows the floating Jacobian to decouple from the nonlinear manifold, introducing systematic biases into the information gain calculation. The percentage of CP95 error absolute difference between them is only 0.47\%. The situation worsens in the L+NG scenario involving non-Gaussian noise. Although the robust kernel suppresses current outliers, the lack of a fixed linearization point prevents this corrected reliability from being stably propagated to subsequent epochs. The most fatal impact occurs in the NL+NG scenario, where heavy nonlinearity demands extreme geometric consistency leading to the CP95 error absolute difference percentage of 1.64\%. Here, an unlocked linearization point causes the Markov chain to undergo cumulative geometric distortion at every frame. 
This demonstrates that the second marginalization is not merely an algebraic requirement but a physical necessity for the survival of recursive filters affected by nonlinear models and non-Gaussian noises.

 \subsection{GNSS/INS integration}
To validate the connection between KFV and SW-FGO in real world environments, we conducted experiments using the data from the UrbanNav dataset \cite{hsu_urbannav_2023} and \cite{hu_pyrtklib_2025}, specifically selecting trajectories through several urban environments of Hong Kong. These scenarios serve a rigorous testbed for both filter-based and optimization-based estimators due to the complex interplay of signal blockages, severe multipath interference, and NLOS receptions. The sensor suite utilized in this study comprises a u-blox F9P dual-frequency GNSS receiver (only the GPS L1 measurement is used), capturing raw measurements at 1 Hz, and an Xsens MTi-10 consumer-grade IMU, which provides 3D accelerations and attitude estimates at 400 Hz via its internal AHRS unit.
To ensure time synchronization and data integrity, all sensor streams were integrated through the Robot Operating System (ROS) \cite{quigley_ros_2009}, with high-frequency IMU data precisely interpolated to align with the GNSS epoch rate. For benchmarking, a centimeter-level ground truth was established using a high-end NovAtel SPAN-CPT system, with the reference trajectory generated via GNSS-RTK/INS tightly coupled forward-backward smoothing using Inertial Explorer 8.9 \cite{novatel_waypoint_2020}, ensuring a high-confidence baseline for error analysis. Following the configuration of our established pipeline \cite{wen_factor_2021}, we maintained consistent parameter settings across all estimators, including motion model noise densities, GNSS weighting strategies, and IMU noise profiles.

\begin{figure}
    \centering
    \includegraphics[width=1\linewidth]{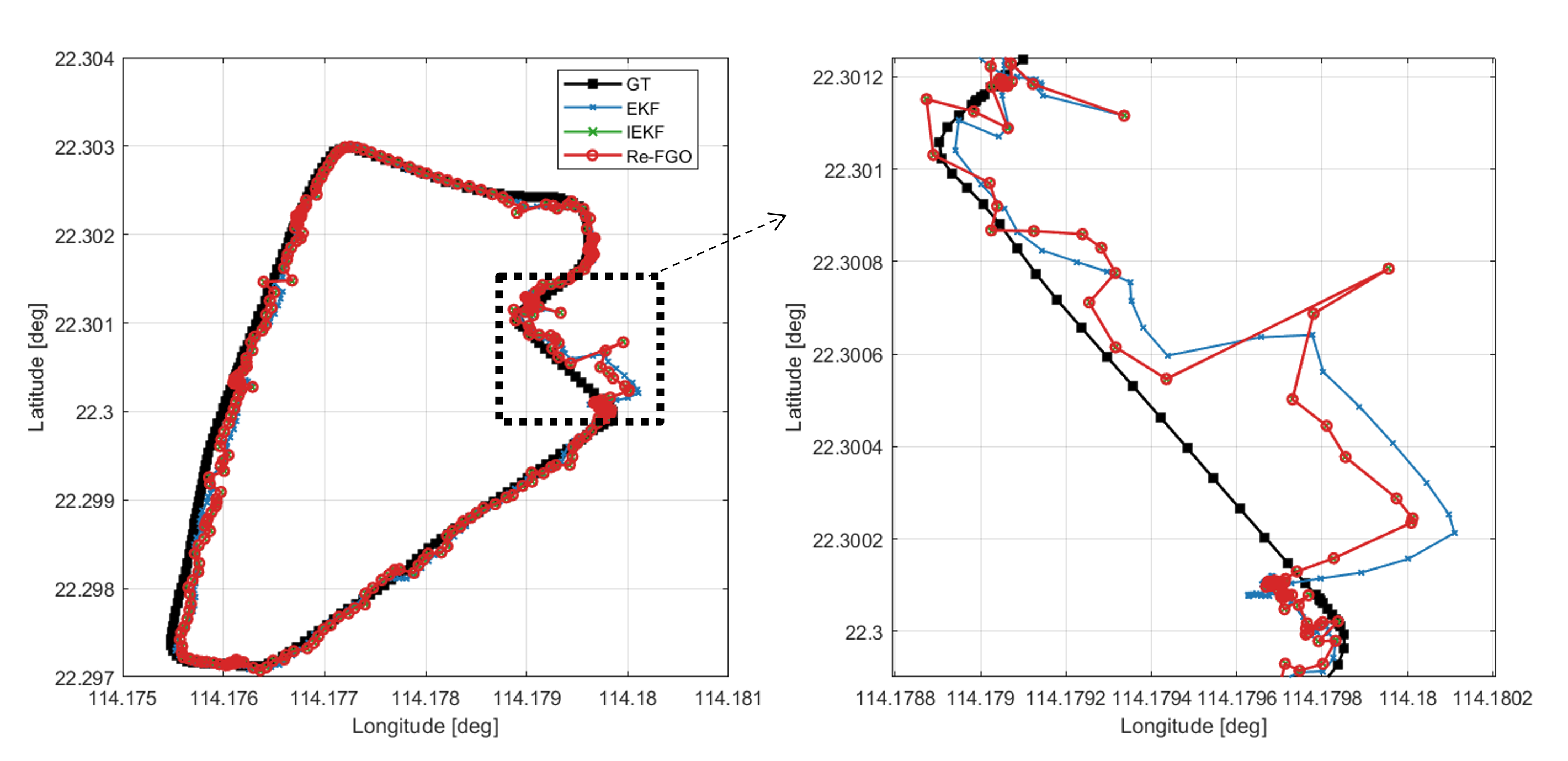}
    \caption{Trajectories of EKF, IEKF and Re-FGO in the medium Urban test. The horizontal positioning performance is compared across the three methods in a realistic urban environment.}
    \label{fig:gins_traj1}
\end{figure}

\begin{figure}
    \centering
    \includegraphics[width=0.7\linewidth]{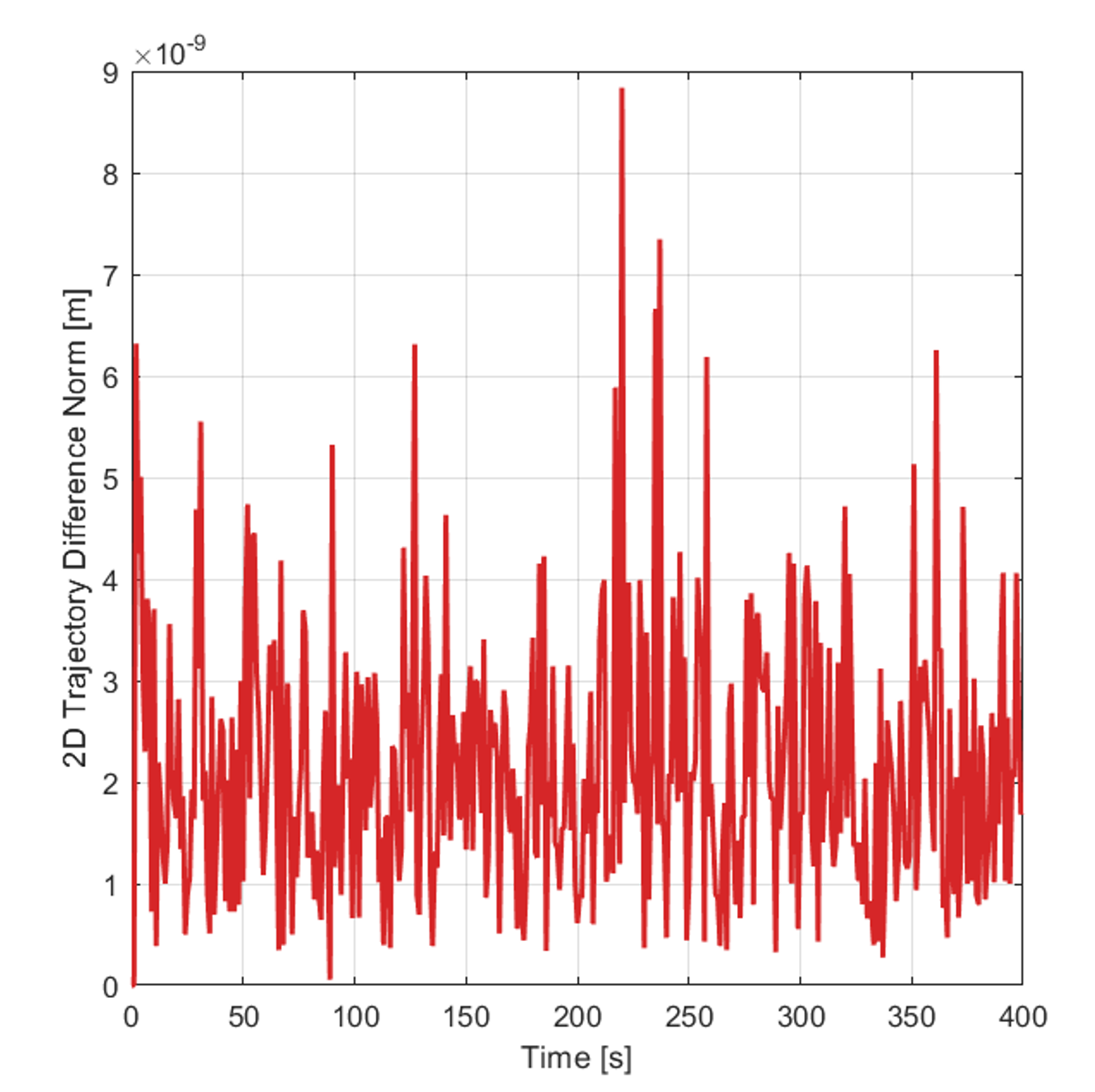}
    \caption{Time series of trajectory difference of IEKF and Re-FGO. The 2-norm of horizontal positions difference is shown for the medium Urban test to validate theoretical equivalence.
}
    \label{fig:gins_diff1}
\end{figure}

\begin{table}[htbp]
\caption{Mean 3D trajectory difference norm between the comparison groups across all scenarios.}
\label{tab:gins_diff1}
\begin{tabular}{@{}ccccc@{}}
\toprule
3D Trajectory Difference Norm [m] & Open    & Medium  & Deep    & Harsh   \\ \midrule
EKF vs. Re-FGO$_1$                & 2.87e-9 & 2.35e-9 & 3.12e-9 & 3.13e-9 \\
IEKF vs. Re-FGO                   & 3.59e-9 & 3.86e-9 & 4.57e-9 & 4.77e-9 \\ \bottomrule
\end{tabular}
\end{table}

 \noindent\textbf{Numerical Equivalence and the Iteration Bridge.} The real-world GNSS/INS integrated trajectories (Fig. \ref{fig:gins_traj1}) and 2D position differences (Fig. \ref{fig:gins_diff1}) demonstrate that the Re-FGO perspective successfully maps the directed logic of recursive filters into the undirected structure of factor graphs. 
The ground truth is marked with black square points. EKF and IEKF are plotted with cross markers while Re-FGO series are plotted with circle markers. Under equivalent conditions, the numerical equivalence is absolute, which is proved by the exact overlay of the trajectories of the IEKF and Re-FGO. 
This reinforces that the Markovian recursive update can be losslessly represented as a specialized, single-epoch factor graph, positioning Re-FGO as an essential structural bridge that maintains the Markov constraint while leveraging the flexibility of the optimization pipeline. 
The overall trends also indicate that increasing the degree of iteration transitions the estimator's behavior from a standard EKF towards that of a full SW-FGO. 
As shown in Table \ref{tab:gins_diff1}, the difference of trajectories is only at the level of $10^{-9}$ m. The real-world GNSS/INS integrated positioning results across multiple scenarios demonstrate that the Re-FGO perspective serves as a structural bridge between recursive filtering and graph optimization. This confirms that the proposed perspective successfully translates the Markovian recursive update into a factor graph representation. 

However, real-world complexity introduces a small challenge where IEKF does not always yield a substantial performance gain over the standard EKF. 
This is primarily because GNSS pseudorange measurements exhibit relatively low nonlinearity due to the vast distance between the receiver and the satellites, making the initial Jacobian already a reasonable approximation. In these scenarios, the primary sources of error are not geometric linearization biases, but rather non-Gaussian noise sources such as multi-path effects and NLOS receptions. Since iterative refinement focuses on minimizing linearization-induced bias rather than suppressing outliers, its effectiveness is inherently limited in real-world GNSS-degraded environments. This finding underscores that while the iteration bridge exists, its practical utility depends on whether the system is limited by model nonlinearity or by the statistical quality of the measurements.

\begin{figure}
    \centering
    \includegraphics[width=1\linewidth]{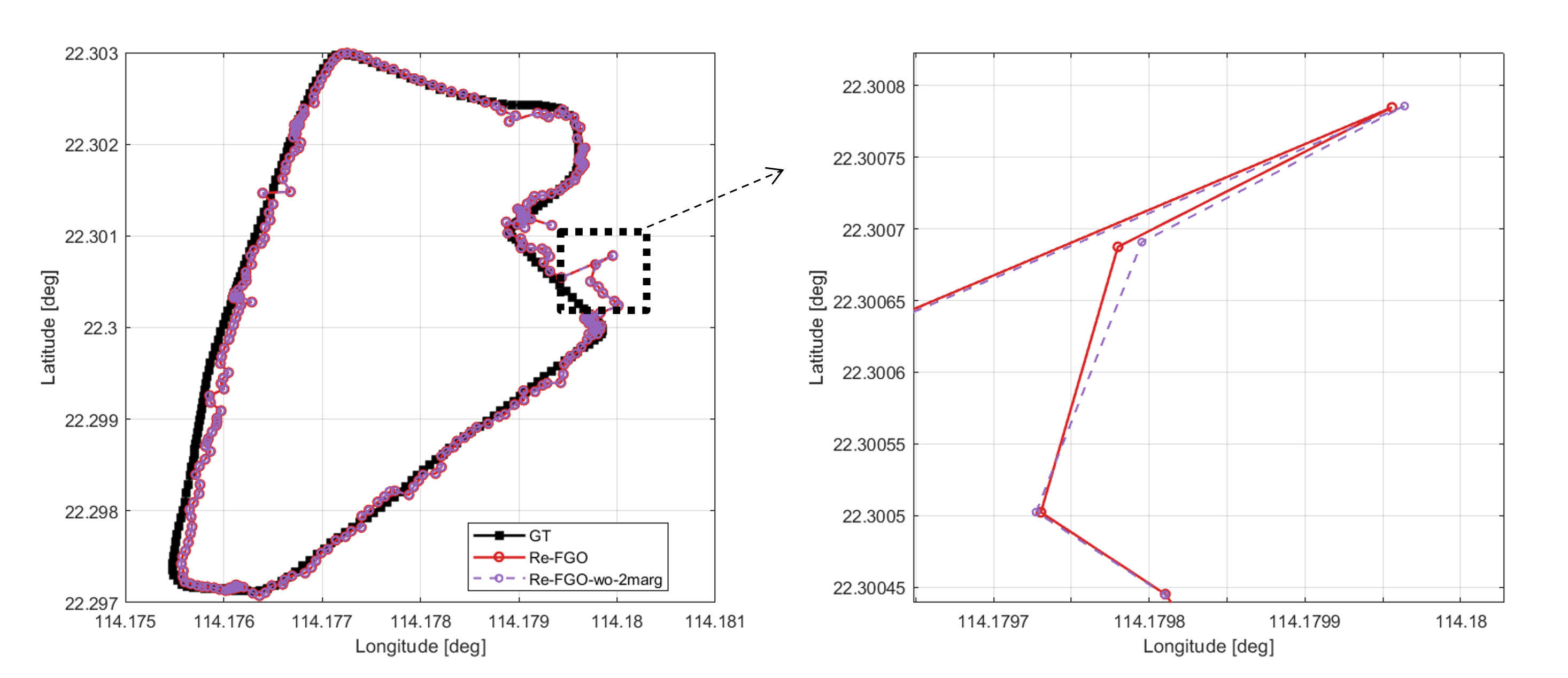}
    \caption{Trajectories of Re-FGO with/without the second marginalization. Comparison between Re-FGO and Re-FGO without the second marginalization in the medium Urban test.}
    \label{fig:gins_traj2}
\end{figure}

\begin{figure}
    \centering
    \includegraphics[width=0.7\linewidth]{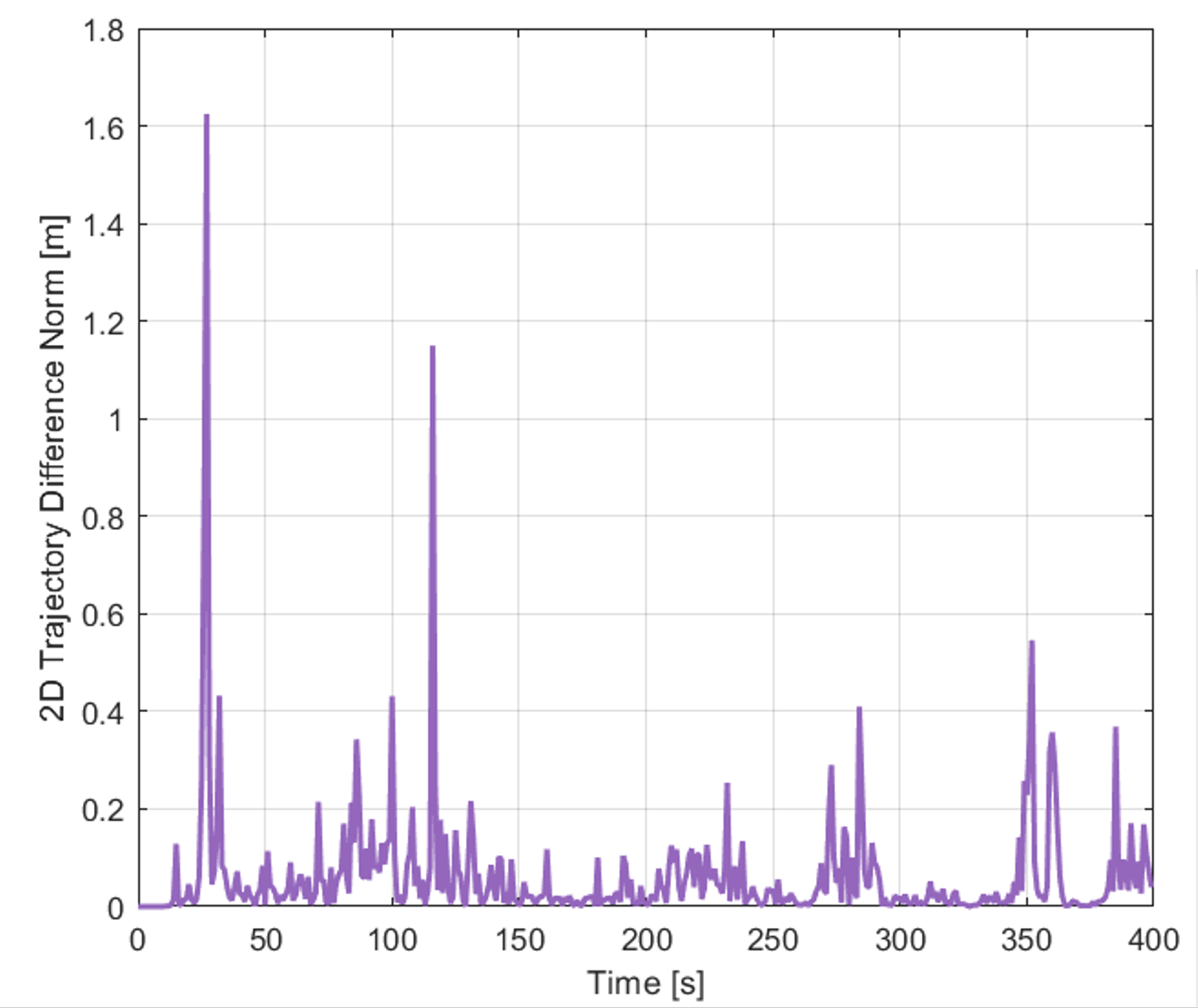}
    \caption{Time series of trajectory difference for marginalization ablation study. The 2-norm of horizontal positions difference between Re-FGO and Re-FGO without second marginalization in the medium Urban test.}
    \label{fig:gins_diff2}
\end{figure}

\noindent\textbf{The Necessity of Marginalization: Fixed vs. Floating Jacobians.} The structural consistency of a recursive estimator is predicated on maintaining a coherent probabilistic link throughout the Markov chain with the mechanism of two-stage marginalization. While the first stage of marginalization serves to compress historical data into a temporal prior and functions analogously to a Schur complement by marginalizing out the previous state, the second stage is specifically tasked with finalizing the linearization point of the current measurement factors. Similar to the simulation, our comparative study between the standard Re-FGO and a variant employing floating Jacobians reveals that failing to anchor these linearization points allows the tangent space of the trajectory to remain alterable. This lack of a fixed gradient results in a fundamental misalignment between the linearized historical prior and the new nonlinear observations, creating a drift in the reference frame that leads to different results, particularly in scenarios characterized by high nonlinearity and non-Gaussian noise.

\begin{table}[htbp]
\centering
\caption{Comparison of CP95 error with and without second marginalization in all the tests.}
\label{tab:marginalization_comparison}
\renewcommand{\arraystretch}{1.2} 
\begin{tabular}{@{}lcccc@{}}
\toprule
CP95 [m]                               & Open    & Medium  & Deep    & Harsh   \\ \midrule
Re-FGO                                 & 9.407   & 17.259  & 18.111  & 38.151  \\
Re-FGO without second marginalization  & 9.414   & 17.282  & 18.096  & 38.553  \\
\textbf{Percentage Difference}         & 0.05\%  & 0.13\%  & 0.08\%  & 1.05\%  \\ \bottomrule
\end{tabular}
\end{table}

To move beyond idealized numerical proofs, it is imperative to validate this mechanism through real-world experiments. Unlike synthetic simulations where noise is often stationary and well-behaved, real-world urban environments present unpredictable signal geometry and complex error profiles that test the limits of estimator stability. We conducted evaluations across four distinct urban canyons in Hong Kong, ranging from relatively open areas to the most densely built-up districts. As shown in Fig. \ref{fig:gins_traj2}, Fig. \ref{fig:gins_diff2} and Table \ref{tab:marginalization_comparison}, our results indicate that the positioning error discrepancy between Re-FGO with and without second-stage marginalization is not uniform and is instead highly sensitive to environmental complexity. In open urban areas, the difference is only 0.05\% and remains relatively modest. However, as the receiver transitions into harsh urban canyons, the error introduced by omitting the second marginalization escalates dramatically, which is denoted by the absolute percentage difference of 1.05\%.

The underlying reason for this environmental sensitivity lies in the coupling between linearization and information within the Markov chain. In dense urban canyons, the GNSS geometry is poor, and measurements are heavily contaminated by multipath/NLOS signals, significantly warping the feasible region. Under these conditions, the iterative optimization aggressively shifts the state estimate to accommodate contradictory measurements. Without the second marginalization to lock the Jacobian, the linearization point is repeatedly re-evaluated at these distorted estimates. This pollutes the historical prior with geometric errors and causes the tangent space to drift away from the true motion manifold. By anchoring the measurement Jacobian, Re-FGO effectively shields the recursive history from re-evaluated distortions, ensuring that corrections are propagated within a consistent geometric reference frame rather than being lost in a drifting search space.

\begin{figure}
    \centering
    \includegraphics[width=1\linewidth]{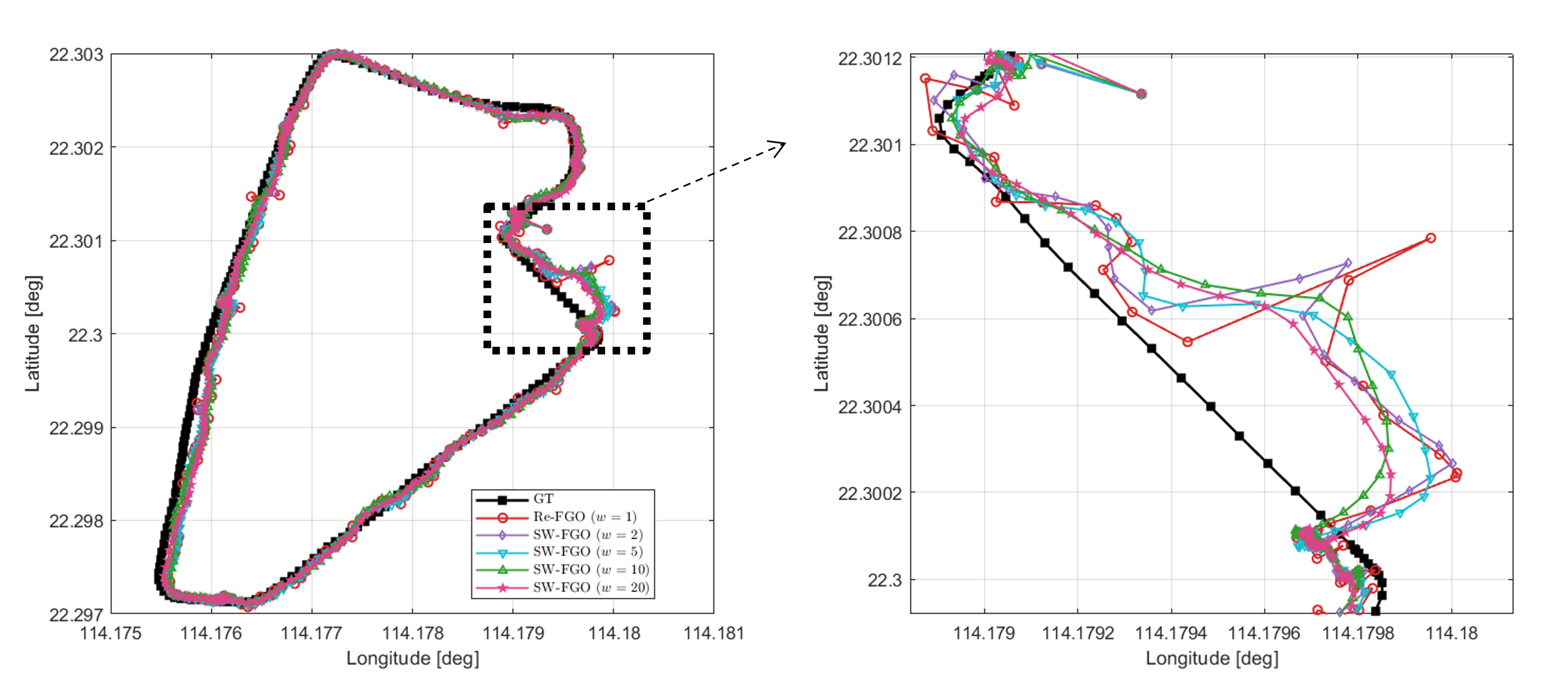}
    \caption{Trajectories of Re-FGO and SW-FGO with different window sizes. Trajectory comparison between Re-FGO ($w=1$) and SW-FGO ($w=2, 5, 10, 20$) in the Medium scenario.}
    \label{fig:gins_traj3}
\end{figure}

\noindent\textbf{Sensitivity Analysis: Scaling from Re-FGO to SW-FGO.} In the previous section, we discussed how iterations help convergence. In this section, we evaluate the performance across various sliding window size from the single-frame Re-FGO ($w=1$) to classical SW-FGO configurations with window sizes $w \in \{2, 5, 10, 20\}$. Under identical convergence thresholds and iteration termination criteria, this scaling analysis acts as a natural bridge between iterative filtering and batch smoothing. 
As shown in the trajectory comparisons of Fig. \ref{fig:gins_traj3}, larger windows produce progressively smoother trajectories. 
With $w=1$ (Re-FGO), the trajectory exhibits sharp spikes and high-frequency tracking jitters because each recursive epoch relies solely on instantaneous, noisy measurements. When we expand the horizon to $w=5$, these fluctuations are visibly reduced. At $w=10$ and $w=20$, the estimated path becomes virtually free of abrupt jumps. This behavior happens because more historical frames act as a temporal buffer. Most sudden outliers or signal disruption in a single frame get effectively averaged out and penalized by the joint constraints of neighboring frames in the factor graph. Thus, a larger sliding window successfully enforces local geometric continuity.

\begin{table}[htbp]
\centering
\caption{Comparison of CP95 error among different estimation methods across the four UrbanNav scenarios.}
\label{tab:gins_cp95_comparison}
\renewcommand{\arraystretch}{1.2}
\begin{tabular}{@{}lcccc@{}}
\toprule
CP95 [m]         & Open   & Medium & Deep   & Harsh  \\ \midrule
EKF              & 11.865 & 17.109 & 18.110 & 46.959 \\
IEKF             & 9.407  & 17.259 & 18.111 & 38.151 \\
Re-FGO           & 9.407  & 17.259 & 18.111 & 38.151 \\
SW-FGO ($w=2$)   & 9.617  & 15.249 & 17.215 & 40.053 \\
SW-FGO ($w=5$)   & 9.406  & 14.582 & 17.021 & 47.082 \\
SW-FGO ($w=10$)  & 10.243 & 16.987 & 17.286 & 55.051 \\
SW-FGO ($w=20$)  & 9.826  & 15.201 & 16.596 & 50.785 \\ \bottomrule
\end{tabular}
\end{table}

However, our empirical data reveals that an increase in window size does not guarantee a certain improvement in absolute positioning accuracy. As reported in Table \ref{tab:gins_cp95_comparison}, the horizontal CP95 errors do not always drop as $w$ expands. Moderate windows ($w=2, 5$) yield modest gains, but the largest window ($w=20$) often fails to outperform $w=2,5$, proving the empirical but not deterministic benefit of larger sliding window.
The underlying reason for this behavior lies in how non-Gaussian noise behaves in urban canyons. In these environments, GNSS errors (like multipath and NLOS) are persistent rather than white and random. While a recursive estimator ($w=1$) immediately marginalizes out the last state, an expanded sliding window inadvertently retains these corruptions inside its active factor chain for a much longer horizon. These errors propagate through the window, resulting in a warped measuring space including outliers. Theoretically, a larger window contains more information, but this capacity to improve absolute accuracy remains latent unless the graph is paired with strong robust cost functions (e.g., Huber or Cauchy) to downweight outliers. Without advanced robust estimation, the accuracy benefit of a larger window is not automatic.

\begin{table}[htbp]
\centering
\caption{Mean computational cost per epoch for the evaluated estimation methods.}
\label{tab:gins_runtime_comparison}
\renewcommand{\arraystretch}{1.2}
\begin{tabular}{@{}lcccc@{}}
\toprule
Runtime per Epoch [ms] & Open   & Medium & Deep   & Harsh  \\ \midrule
EKF                    & 0.44   & 0.43   & 0.44   & 0.45   \\
IEKF                   & 0.75   & 0.71   & 0.75   & 0.72   \\
Re-FGO                 & 5.80   & 6.18   & 8.25   & 14.20  \\
SW-FGO ($w=2$)         & 9.78   & 10.06  & 24.89  & 12.80  \\
SW-FGO ($w=5$)         & 23.52  & 22.30  & 59.79  & 28.83  \\
SW-FGO ($w=10$)        & 48.31  & 44.50  & 118.93 & 45.40  \\
SW-FGO ($w=20$)        & 130.94 & 132.13 & 287.03 & 115.10 \\ \bottomrule
\end{tabular}
\end{table}

Moreover, one trend that never fails is that larger windows cost much more computation. As summarized in Table \ref{tab:gins_runtime_comparison}, the mean runtime per epoch generally increases with larger window sizes. For example, in the medium test, the mean runtime escalates from $6.18 \text{ ms}$ ($w=1$) and $10.06\text{ ms}$ ($w=2$) to $132.13\text{ ms}$ ($w=20$). This result stems from the cubic complexity of factorizing a denser Hessian matrix. Within this range, Re-FGO ($w=1$) corresponds to the smallest window size, conceptually aligning with the IEKF, whereas larger $w$ approach batch smoothing. Hence, the runtime data positions Re-FGO as a natural bridge between IEKF and SW-FGO, with its computational cost lying between these two methods. Interestingly, an anomaly occurs in the harsh scenario, where Re-FGO ($w=1$) requires a slightly longer runtime ($14.20\text{ ms}$) compared to SW-FGO ($w=2$) ($12.80\text{ ms}$). This behavior stems from the severe lack of reliable observations in the harsh urban canyon. For a minimal window size of $w=1$, the structural sparsity leaves the joint Hessian ill-conditioned, forcing the Gauss-Newton solver to undergo more iterations or line-search steps to reach numerical convergence. Conversely, by expanding the window to $w=2$, the historical state trajectory introduces implicit spatial regularizations that significantly improve the matrix condition number, thereby accelerating the linear system factorization and reducing the overall optimization runtime.

\backmatter

\bmhead{Supplementary information}
The Supplementary Information accompanying this manuscript contains detailed algebraic proofs of the Re-FGO and IEKF equivalence (Supplementary Notes 1 and 2).




\section*{Data Availability}
The datasets generated and/or analysed during the current study are available at \url{https://github.com/Baoshan-Song/KFV-FGO-Comparison}.

\section*{Code Availability}
The open-sourced data and codes are available at \url{https://github.com/Baoshan-Song/KFV-FGO-Comparison}.

\section*{Acknowledgements}
The authors sincerely thank Yuan Li, Zhengdao Li, Yihan Zhong, Xiao Xia, Penggao Yan, Penghui Xu, and Hongmin Zhang for their valuable suggestions.


\bibliography{npjwt_bib}

\setcounter{section}{0}

\section{Supplementary Note 1}\label{sec:preliminary}

Both EKF and SW-FGO are rooted in MAP estimation principle. Given a sequence of observations $\mathcal{Z} = \{z_1, \dots, z_k\}$ and states $\mathcal{X} = \{x_0, x_1, \dots, x_k\}$, the MAP estimate is defined as: 
\begin{equation} \label{equ:bayes_appendix} 
\mathcal{X}^* = \arg\max_{\mathcal{X}} p(\mathcal{Z} \mid \mathcal{X}) p(\mathcal{X}) = \arg\min_{\mathcal{X}} \left( -\log p(\mathcal{Z} \mid \mathcal{X}) - \log p(\mathcal{X}) \right). 
\end{equation}
where $p(\mathcal{X} \mid \mathcal{Z})$ is the posterior probability; $p(\mathcal{Z} \mid \mathcal{X}) $ and $p(\mathcal{X})$ are the likelihood and prior probability separately. In SW-FGO, the estimate is typically obtained by joint optimization over all states within the sliding window simultaneously: 
\begin{equation}
    p(x_{1:k}|z_{1:k}).
\end{equation}
In contrast, the EKF enforces a first-order Markov assumption to realize a recursive Bayesian inference: 
\begin{equation} p(x_{k}|z_{1:k})\propto \underbrace{p(z_k|x_k)}_{\text{Likelihood}} \int \underbrace{p(x_k|x_{k-1})p(x_{k-1}|z_{1:k-1})}_{\text{Prediction}} dx_{k-1} 
\end{equation}

\section{Supplementary Note 2}
\label{sec:proof_prop1}
In this section, we prove Proposition 1 by demonstrating that the two-stage marginalization protocol in Re-FGO degenerates the undirected factor graph optimization into the recursive closed-form updates of the IEKF.

\subsection{Stage 1: State Elimination as IEKF Prediction}
Following the Markov assumption, the estimation at epoch $k$ starts with the joint posterior of the previous state and the current state given past measurements. In the factor graph, this is represented by the WLS problem over the expanded state vector $\mathbf{x}^- = [x_{k-1}^T, x_k^T]^T$:
\begin{equation} 
\label{equ:stage1_wls}\arg\min_{x_{k-1}, x_k} \left( \left\| x_{k}-f(x_{k-1})\right\|^2_{\mathbf{Q}_k}+\left\| x_{k-1}-{x}^+_{k-1}\right\|^2_{\mathbf{P}_{k-1}^+} \right)
\end{equation}
Linearizing the motion model $f(x_{k-1})$ around the previous posterior $x_{k-1}^+$ using $\mathbf{F}_{k-1} = \frac{\partial f}{\partial x}|_{x_{k-1}^+}$, we define the Jacobian $\mathbf{J}^-$ and the information matrix $\mathbf{\Lambda}^-$ of the joint system:
\begin{equation}
\mathbf{\Lambda}^- = {\mathbf{J}^-}^T {\mathbf{W}^-}^{-1} {\mathbf{J}^-} = \begin{bmatrix}{\mathbf{P}_{k-1}^+}^{-1} + \mathbf{F}_{k-1}^T \mathbf{Q}^{-1} \mathbf{F}_{k-1} & -\mathbf{F}_{k-1}^T \mathbf{Q}^{-1} \\ -\mathbf{Q}^{-1} \mathbf{F}_{k-1} & \mathbf{Q}^{-1}\end{bmatrix}
\end{equation}
To maintain a window size of $w=1$, we marginalize out the previous state $x_{k-1}$ using the Schur complement. The resulting marginalized Hessian $\mathbf{H}_{k}^-$ represents the inverse of the predictive covariance:
\begin{equation}
\label{equ:schur_marg}
\mathbf{H}_{k}^- = \mathbf{Q}^{-1} - (-\mathbf{Q}^{-1} \mathbf{F}_{k-1}) \left( {\mathbf{P}_{k-1}^+}^{-1} + \mathbf{F}_{k-1}^T \mathbf{Q}^{-1} \mathbf{F}_{k-1} \right)^{-1} (-\mathbf{F}_{k-1}^T \mathbf{Q}^{-1})
\end{equation}
By applying the matrix inversion lemma to \eqref{equ:schur_marg}, we obtain:
\begin{equation}
\mathbf{H}_{k}^- = (\mathbf{Q} + \mathbf{F}_{k-1} \mathbf{P}_{k-1}^+ \mathbf{F}_{k-1}^T)^{-1} 
\implies
\mathbf{P}_k^- = \mathbf{F}_{k-1} \mathbf{P}_{k-1}^+ \mathbf{F}_{k-1}^T + \mathbf{Q}
\end{equation}
This confirms that the first stage of marginalization in Re-FGO is algebraically identical to the IEKF prediction step, producing the predictive prior $\mathcal{N}(x_k^-, \mathbf{P}_k^-)$ where $x_k^- = f(x_{k-1}^+)$.

\subsection{Stage 2: Factor-only Anchoring as IEKF Update}

After obtaining the prior factor from Stage 1, we incorporate the measurement factor $z_k$. In IEKF, the update is an iterative process. Within the Re-FGO framework, this corresponds to Gauss-Newton iterations on the cost function:
\begin{equation}\label{equ:iekf_update_app}\arg\min_{\delta x_{k,j}} \left( \left\| z_k - h(x_{k,j}) - \mathbf{H}_{k,j}\delta x_{k,j} \right\|^2_{\mathbf{R}^{}} + \left\| x_{k,j} + \delta x_{k,j} - x_k^- \right\|^2_{\mathbf{P}_{k}^-} \right)
\end{equation}
where $j$ is the iteration index and $\delta x_{k,j}$ is the update increment. Solving for $\delta x_{k,j}$ yields:
\begin{equation}
( {\mathbf{P}_{k}^-}^{-1} + \mathbf{H}_{k,j}^T \mathbf{R}^{-1} \mathbf{H}_{k,j} ) \delta x_{k,j} = \mathbf{H}_{k,j}^T \mathbf{R}^{-1} (z_k - h(x_{k,j})) - {\mathbf{P}_{k}^-}^{-1} (x_{k,j} - x_k^-)
\end{equation}
By defining the Kalman Gain $\mathbf{K}_{k,j}$ as $\mathbf{P}_k^- \mathbf{H}_{k,j}^T (\mathbf{H}_{k,j} \mathbf{P}_k^- \mathbf{H}_{k,j}^T + \mathbf{R})^{-1}$, the iterative solution for the state update is:
\begin{equation}
x_{k,j+1} = x_{k,j} + \delta x_{k,j} = x_k^- + \mathbf{K}_{k,j} (z_k - h(x_{k,j}) - \mathbf{H}_{k,j}(x_k^- - x_{k,j}))
\end{equation}
This is the standard IEKF iterative update. 

To finalize the recursive cycle, Re-FGO performs factor-only marginalization. Unlike standard FGO which may relinearize historical factors, Re-FGO anchors the measurement Jacobian $\mathbf{H}_{k,j}$ at the final converged MAP estimate $x_k^+$. This "locks" the linearization point, ensuring that the information condensed into the next epoch's prior is consistent with the filter's update logic, thereby preventing relinearization drift and satisfying the Markov assumption.
After reaching the posterior mode $\mathbf{x}_k^*$ through iterative Gauss-Newton steps, Re-FGO executes the second stage of marginalization to anchor the measurement information. To prove the equivalence to the IEKF update, we examine the system's linear alignment. At the $j$-th (final) iteration, the WLS objective is composed of two factors: the predictive prior from Stage 1 and the linearized measurement. We define the incremental error-state equation as:
\begin{equation}
\mathbf{r}(\delta \mathbf{x}) = \mathbf{A} \delta \mathbf{x} - \mathbf{b}
\end{equation}
where the combined Jacobian matrix $\mathbf{A}$ and the residual vector $\mathbf{b}$ are constructed as:
\begin{equation}
\mathbf{A} = 
\begin{bmatrix} 
(\mathbf{P}_k^-)^{-1/2} & \mathbf{R}^{-1/2} \mathbf{H}_k \end{bmatrix}, \quad
\mathbf{b} = \begin{bmatrix} (\mathbf{P}_k^-)^{-1/2} ( \mathbf{x}_k^- - \mathbf{x}_k^* ) & \mathbf{R}^{-1/2} ( \mathbf{z}_k - h(\mathbf{x}_k^*) ) 
\end{bmatrix}
\end{equation}
Re-FGO applies a Thin QR factorization to the joint Jacobian $\mathbf{A}$ to eliminate the state dependency while preserving the information structure:
\begin{equation}
\mathbf{Q}^T \mathbf{A} = 
\begin{bmatrix} \mathbf{R}_{qr} & \mathbf{0} \end{bmatrix}, 
\quad \mathbf{Q}^T \mathbf{b} = \begin{bmatrix} \mathbf{d} & \mathbf{e} \end{bmatrix}
\end{equation}
where $\mathbf{R}_{qr}$ is an upper-triangular matrix. Since $\mathbf{Q}$ is an orthogonal matrix, the Hessian of the factor graph $\mathbf{H}_{fgo}$ satisfies:\begin{equation}\mathbf{H}_{fgo} = \mathbf{A}^T \mathbf{A} = (\mathbf{Q} \mathbf{R}_{qr})^T (\mathbf{Q} \mathbf{R}_{qr}) = \mathbf{R}_{qr}^T \mathbf{R}_{qr}
\end{equation}
Expanding $\mathbf{A}^T \mathbf{A}$ directly gives:
\begin{equation}
\mathbf{H}_{fgo} = (\mathbf{P}_k^-)^{-1} + \mathbf{H}_k^T \mathbf{R}^{-1} \mathbf{H}_k
\end{equation}
In the IEKF framework, the posterior covariance $\mathbf{P}_k^+$ is updated via the information form:
\begin{equation} \label{equ:iekf_info_form}
(\mathbf{P}_k^+)^{-1} = (\mathbf{P}_k^-)^{-1} + \mathbf{H}_k^T \mathbf{R}^{-1} \mathbf{H}_k
\end{equation}
By comparing these terms, it is evident that $\mathbf{R}_{qr}^T \mathbf{R}_{qr} = (\mathbf{P}_k^+)^{-1}$. This demonstrates that the upper-triangular matrix $\mathbf{R}_{qr}$ produced by Re-FGO is the Cholesky factor of the IEKF posterior information matrix.

To satisfy the Markov assumption, Re-FGO solves the anchored linear system 
$\mathbf{R}_{qr} \delta \mathbf{x}^* = \mathbf{d}$
for the final state increment. Expanding the normal equation 
$\mathbf{A}^T \mathbf{A} \delta \mathbf{x}^* = \mathbf{A}^T \mathbf{b}$ yields:
\begin{equation}
\left( (\mathbf{P}_k^-)^{-1} + \mathbf{H}_k^T \mathbf{R}^{-1} \mathbf{H}_k \right) \delta \mathbf{x}^* = \mathbf{H}_k^T \mathbf{R}^{-1} ( \mathbf{z}_k - h(\mathbf{x}_k^* )) + (\mathbf{P}_k^-)^{-1} ( \mathbf{x}_k^- - \mathbf{x}_k^*)
\end{equation}
Using the Matrix Inversion Lemma and defining the Kalman Gain 
$\mathbf{K}_k = \mathbf{P}_k^- \mathbf{H}_k^T ( \mathbf{H}_k \mathbf{P}_k^- \mathbf{H}_k^T + \mathbf{R} )^{-1}$,
the solution for $\delta \mathbf{x}^*$ is:
\begin{equation}\delta \mathbf{x}^* = \mathbf{K}_k ( \mathbf{z}_k - h(\mathbf{x}_k^*) ) + ( \mathbf{I} - \mathbf{K}_k \mathbf{H}_k ) ( \mathbf{x}_k^- - \mathbf{x}_k^* )\end{equation}
which is exactly the iterative update step of the IEKF.

\subsection{Pipeline of Re-FGO}

The mathematical isomorphism between Re-FGO and the IEKF is established by mapping the two-stage marginalization protocol directly onto the recursive Bayesian inference cycle. The main steps are concluded in Table \ref{tab:algorithm_comparison}. In the first stage, the elimination of the previous state $\mathbf{x}_{k-1}$ via the Schur complement on the joint motion-prior Hessian produces a marginalized predictive prior; by applying the matrix inversion lemma, it is proven that the resulting information matrix $\mathbf{H}_k^-$ is algebraically identical to the inverse of the IEKF predictive covariance $\mathbf{P}_k^- = \mathbf{F}_{k-1} \mathbf{P}_{k-1}^+ \mathbf{F}_{k-1}^\top + \mathbf{Q}$. In the second stage, the iterative Gauss-Newton optimization over the current state $\mathbf{x}_k$ replicates the IEKF’s iterative refinement. Upon convergence at $\mathbf{x}_k^*$, the thin QR factorization of the joint prior-measurement Jacobian yields an upper-triangular factor $\mathbf{R}_{qr}$ such that $\mathbf{R}_{qr}^\top \mathbf{R}_{qr}$ exactly recovers the IEKF posterior information matrix $(\mathbf{P}_k^+)^{-1}$. Solving the system $\mathbf{R}_{qr} \delta\mathbf{x} = \mathbf{d}$ yields a state increment $\delta\mathbf{x}^*$ identical to the IEKF update, ensuring that the mean and covariance propagated to the next epoch are numerically equivalent in both frameworks. 

Beyond numerical equivalence, the structural alignment provided by Re-FGO enforces the Markov property through a unique anchoring mechanism that bridges the gap between fluid sliding windows and recursive filters. While standard SW-FGO allows for the retrospective relinearization of all states within its window, Re-FGO’s second marginalization phase effectively "locks" the measurement Jacobian at the final posterior mode $\mathbf{x}_k^*$. By discarding the orthogonal residual $\mathbf{e}$ and storing only the anchored prior $(\mathbf{R}_{qr}, \mathbf{d})$, the framework ensures that the measurement information is irreversibly condensed before being passed to the next epoch. This prevents the backward flow of information and relinearization drift, satisfying the strict conditional independence required by a Markov chain. Consequently, by applying Markov assumption and this two-stage marginalization, the undirected graph optimization of SW-FGO degenerates to the closed-form recursion of the IEKF, providing a theoretical bridge that preserves the iterative robustness of factor graphs within a computationally efficient filter structure. Proposition 1 is thus proven.

\end{document}